\documentclass{article}
\PassOptionsToPackage{table}{xcolor}
\usepackage{fontspec}
\AtBeginDvi{}
\usepackage{geomerge_preprint,times}
\usepackage{amsmath,amssymb,mathtools,amsthm}
\usepackage{graphicx,booktabs,array,needspace}
\usepackage{xcolor}
\usepackage{geomerge_appendix_style}
\usepackage{xurl,hyperref}
\hypersetup{pdftitle={GeoMerge-OPD: Balanced Skill Integration in Vision-Language-Action Models},pdfauthor={Le Yang, Haoyang Lyu, Honglei Jin, Wenshuo Chen, Jiawei Liang, Wenqiang Wang, Yutao Yue, Xiaochun Cao},pdfsubject={},pdfkeywords={},colorlinks=true,citecolor=blue,linkcolor=blue,urlcolor=blue}

\usepackage{tabularx}
\definecolor{MTblue}{HTML}{E5F4FC}
\definecolor{MTgray}{RGB}{226,226,226}
\newcommand{\MTsetup}{\setlength{\abovecaptionskip}{4pt}\fontsize{9}{10.4}\selectfont\setlength{\tabcolsep}{3.5pt}\renewcommand{\arraystretch}{1.05}}
\newcommand{\MTgroup}[2]{\addlinespace[1pt]\multicolumn{#1}{@{}l}{\textcolor{black}{\textbf{#2}}}\\[.4pt]}
\newcommand{\MTnote}[1]{\par\vspace{1pt}\begin{minipage}{\linewidth}\fontsize{8}{9}\selectfont #1\end{minipage}}

\usepackage{wrapfig}

\usepackage{capt-of}

\usepackage{placeins}

\title{GeoMerge-OPD: Balanced Skill Integration\\
in Vision-Language-Action Models}
\author{%
  \textbf{Le Yang}\textsuperscript{1,*}\quad \textbf{Haoyang Lyu}\textsuperscript{2,*}\quad \textbf{Honglei Jin}\textsuperscript{3,*}\quad \textbf{Wenshuo Chen}\textsuperscript{3,*,\textdagger}\\[2pt]
  \textbf{Jiawei Liang}\textsuperscript{1}\quad \textbf{Wenqiang Wang}\textsuperscript{1}\quad \textbf{Yutao Yue}\textsuperscript{3,\textdaggerdbl}\quad \textbf{Xiaochun Cao}\textsuperscript{1,\textdaggerdbl}\\[4pt]
  {\small \textsuperscript{1}\,Sun Yat-sen University\qquad \textsuperscript{2}\,City University of Hong Kong}\\
  {\small \textsuperscript{3}\,The Hong Kong University of Science and Technology (Guangzhou)}\\[3pt]
  {\footnotesize \textsuperscript{*}\,Co-first authors (equal contribution).
  \quad \textsuperscript{\textdagger}\,Project leader.
  \quad \textsuperscript{\textdaggerdbl}\,Co-corresponding authors.}%
}
\begin{document}
\maketitle
\begin{abstract}
Vision-language-action (VLA) models excel at individual robotic skills, yet integrating heterogeneous skills into a single policy remains a key challenge for general-purpose robotic control, especially in whole-body tasks with differing manipulation and locomotion requirements. Common multi-skill integration strategies, such as joint fine-tuning and parameter merging, can induce unintended actions during closed-loop execution, severely degrading individual skills. Task-conditioned analysis of expert updates reveals response energy concentrated in a few dominant directions and opposing expert responses along high-energy directions within a shared, task-relevant response subspace. This structure co-occurs with unintended action shifts after merging. We propose GeoMerge-OPD, combining response-space regularization, parameter merging, and on-policy distillation. During independent expert fine-tuning, G-isometry constrains update-response gains within task-activation subspaces; the expert parameters are then statically merged into a student with preliminary multitask capabilities. Fine-tuned task experts serve as teachers, providing velocity-field supervision at the environment observations and intermediate action-flow states visited by the student to calibrate residual behavioral deviations. Experiments with two VLA backbones show strong, balanced humanoid skill integration on SIMPLE and consistent performance on LIBERO and LIBERO-PRO, outperforming joint fine-tuning and multitask reinforcement-learning baselines while maintaining robustness under perturbations.

\end{abstract}
\section{Introduction}
Vision-language-action (VLA) models adapt pretrained vision-language models to robotic control using demonstration data, combining semantic understanding and referential grounding with strong performance on individual manipulation tasks \citep{knowledge_insulation,pi05,actions_as_language}. Generalist robots, however, must integrate multiple skills within one policy; humanoids must additionally coordinate upper-body manipulation and lower-body locomotion \citep{wholebodyvla,psi0}. This makes reliable integration of heterogeneous skills a central challenge for VLA models.

A common approach is joint fine-tuning on demonstrations pooled across skills. Although effective for homogeneous tasks, it can improve some skills at the expense of others when task demands differ \citep{openvla_oft,rethink_vla_scaling}. These trade-offs also arise with a shared control interface: with a \(\pi_{0.5}\) backbone, joint training on a stream of real-robot tasks improved one normalized task score from 52 to 92 while reducing another from 100 to 33.3 \citep{continualvla}. On whole-body tasks, jointly fine-tuned \(\Psi_0\) underperformed its single-task counterparts across all evaluated tasks, with some skills nearly lost \citep{psi0}.

An alternative is to fine-tune experts separately and merge their parameters, an approach effective in language and vision models \citep{ties,regmeanpp,isoc}. MergeVLA extends this approach to VLAs through task masks, dedicated action heads, and test-time routing \citep{mergevla}. However, we find that directly merging whole-body VLA experts can severely degrade individual skills, even when the task-specific experts retain those skills. In our \(\Psi_0\) experiments, the merged policy issued unintended locomotion commands during BendPick, a task requiring bending and grasping in place, resulting in task failure. These failures motivate our central question: how can heterogeneous task behaviors be retained in shared parameters to form a balanced, robust policy without maintaining separate models for individual tasks?

Prior work documents concentrated update spectra and interference arising from disagreements in parameter signs or singular directions \citep{ties,tsvmerge,isoc,dcmerge}. It also shows that the functional effects of updates depend on inputs and may concentrate in low-dimensional activation or output-shift subspaces \citep{wudi,esm}. Building on these findings, we analyze task-conditioned responses of \(\Psi_0\) experts, using task activation statistics to characterize the functional effects of their parameter updates. We observe opposing components within a shared task-relevant response subspace, concentrated along high-energy response directions. This structure co-occurs with unintended action shifts after merging, suggesting that energy concentration can amplify the impact of opposing responses. These observations motivate regularizing update responses before merging and calibrating the resulting policy afterward.

We propose GeoMerge-OPD, which combines regularized expert fine-tuning, static parameter merging, and on-policy distillation (OPD). Starting from a common pretrained backbone, each expert is fine-tuned with G-isometry to constrain update-response gains in task-activation subspaces. This regularizer acts on responses induced by task inputs, reducing excessive dominance by a few directions. Static merging then establishes a balanced initialization with preliminary multitask competence. Inspired by multi-teacher OPD for integrating heterogeneous expert capabilities in language and text-to-image models \citep{mopd,flowopd}, we use OPD to consolidate skills from task-specific VLA experts into a unified student policy. The student interacts with the environment in closed loop and generates its own action-flow trajectories. The corresponding task expert provides dense velocity-field supervision at the environment observations and intermediate flow states actually visited by the student. This supervision calibrates residual behavioral deviations under the student's own state and action-generation distributions, further coordinating multitask skills within the shared model.

We evaluate GeoMerge-OPD on LIBERO \citep{libero}, LIBERO-PRO \citep{liberopro}, and SIMPLE \citep{simple}, using \(\pi_{0.5}\) and \(\Psi_0\) as the two VLA backbones, covering tabletop manipulation, whole-body loco-manipulation, and generalization to perturbed evaluation conditions. Our main contributions are:

\begin{enumerate}
\item \textbf{Empirical diagnosis of multitask VLA integration.} We characterize failures in merging task-specific VLA experts and identify concentrated update responses and opposing components in shared task-relevant subspaces, connecting these observations to unintended behavioral shifts in closed-loop execution.

\item \textbf{A framework for balanced skill consolidation.} GeoMerge-OPD combines G-isometry regularization and static merging to establish a balanced initialization, followed by task-routed OPD to calibrate residual behavior on student-visited states and action-flow trajectories, consolidating heterogeneous skills in shared parameters.

\item \textbf{Skill consolidation and robustness across two backbones.} GeoMerge-OPD achieves 26/30 whole-body successes on SIMPLE versus 22/30 for Joint FT from the same G-regularized merge, reaches 97.1\% on LIBERO versus 96.2\% for Joint FT from the pretrained base, and attains the highest average LIBERO-PRO success (67.3\%) across five perturbation axes among the evaluated policies.
\end{enumerate}

\section{Related Work}
\subsection{Model Merging and VLA Skill Integration}

Model merging research has expanded from parameter combination to conflict resolution, task-subspace modeling, and spectral regulation to improve the integration of expert capabilities \citep{ties,tsvmerge,isoc}.
Mergeability has also been incorporated into expert training and architectural design to prepare models for subsequent consolidation \citep{mergopt,mergevla}.
In parallel, studies of input-dependent update effects use input or feature subspaces to characterize how parameter updates influence task behavior \citep{wudi,esm}.
Together, these studies provide a basis for understanding post-merging capability retention through both expert-update structure and task-dependent functional responses, informing the analysis of behavioral degradation in heterogeneous robotic skill integration.

\subsection{Policy Distillation and Multi-Teacher Capability Integration}

Behavioral data generated by specialized experts can supervise generalist robotic policies, consolidating experience across tasks or manipulated objects within a single policy \citep{rldg,unigrasp_transformer}.
Applying teacher supervision to student-generated sequences or trajectories helps mitigate distribution mismatch between training and execution and correct errors induced by the student's behavior \citep{gkd,vlaopd}.
In large language models and text-to-image generation, multi-teacher OPD further integrates heterogeneous capabilities into a unified student through supervision from domain-specialized teachers on student-generated trajectories \citep{mopd,flowopd}.
To our knowledge, multi-teacher OPD for integrating heterogeneous skills across VLA tasks remains underexplored.
Inspired by this capability integration paradigm, we use task-specific VLA teachers to provide dense velocity-field supervision along the student's own action-flow trajectories, conditioned on observations encountered during its environment rollouts, to coordinate multi-task skills after parameter merging.

\section{Why Multi-Task Integration Fails}
\subsection{Capability collapse after integration}

We refer to severe degradation of individual skills after multitask integration as capability collapse. On the BendPick, CloseDoor, and OpenOven tasks with \(\Psi_0\), both joint fine-tuning and static parameter merging exhibited this phenomenon, with BendPick nearly failing completely. An examination of failed episodes showed that the merged policy continuously issued lower-body movement commands in BendPick scenes requiring bending and grasping in place, whereas the corresponding task-specific expert did not issue such commands. This unintended movement caused the merged policy to deviate from the control behavior required by the task.

\subsection{Response energy concentration and opposing responses}

\begin{wrapfigure}[17]{R}{0.52\textwidth}
\setlength{\abovecaptionskip}{5pt}
\centering
\includegraphics[width=\linewidth,trim={110bp 190bp 115bp 125bp},clip]{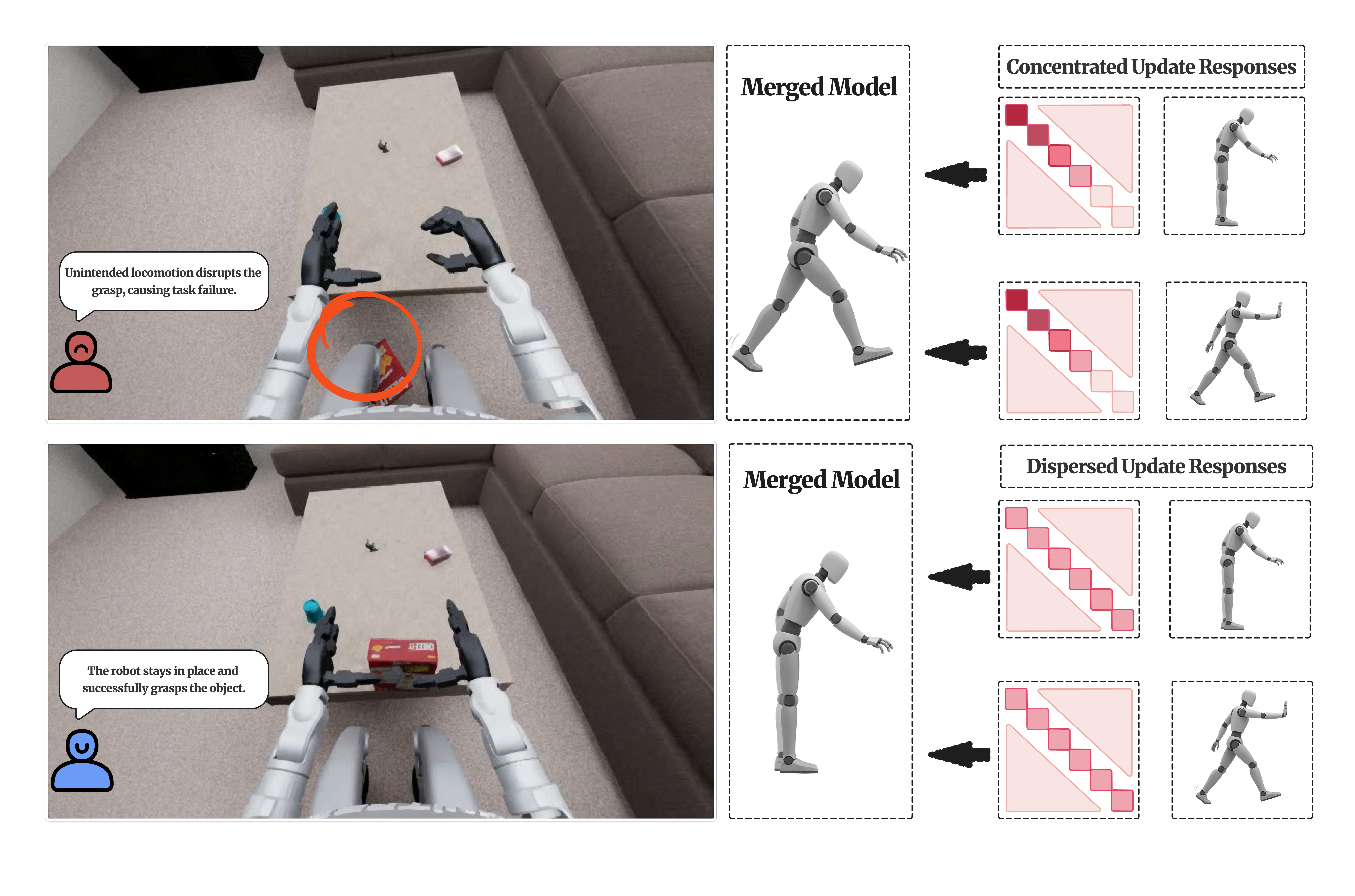}
\caption{\small BendPick after merging: unintended locomotion (top) and stationary grasping with regularized experts (bottom). Response matrices are schematic.}
\label{fig:qualitative}
\end{wrapfigure}

To analyze this deviation, we examine the responses induced by expert updates on task inputs. For a fixed layer, let $G_t=\mathbb E_t[xx^\top]$ denote the input second moment of task $t$. The response energy induced by any weight change $A$ is

\begin{equation}
\mathcal E_t(A)=\mathbb E_t\|Ax\|_2^2
=\|AG_t^{1/2}\|_F^2.
\label{eq:response-energy}
\end{equation}

For an expert update $\Delta W_t=W_t-W_0$, the squared singular values of $\Delta W_tG_t^{1/2}$ describe the energy along the principal response directions. If most energy concentrates in a few directions, deviations along those directions affect a large fraction of the task response. When other experts introduce opposing components along these directions, parameter merging may cancel the adjustments required by the task or allow the other experts' responses to dominate, causing the merged result to deviate from task requirements. Appendix~\ref{app:merge-response} analyzes this behavior under fixed activations. Degradation during joint fine-tuning also depends on task gradients at the current parameters, with the relevant conditions provided in Appendix~\ref{app:joint-training}.

We examine the update responses of task-specific fine-tuned experts on \(\Psi_0\), analyzing their energy distributions and directional relationships across experts (Appendix~\ref{app:mechanism-diagnostics}). For the standard fine-tuned CloseDoor and BendPick experts, the top five principal response directions account for 93\% and 78\% of response energy, respectively. Within the principal response subspace defined with BendPick as the reference, the median directional cosine between the two experts' responses is $-0.73$, indicating a pronounced opposing relationship. The tasks also differ in their lower-body control requirements: CloseDoor requires movement, whereas BendPick requires remaining stationary. This response structure co-occurs with capability collapse on BendPick after merging.

To examine the connection between response structure and behavioral degradation, we constrain the update responses within task-activation subspaces during fine-tuning. With this constraint, the median energy shares of the top five directions for CloseDoor and BendPick are 14.3\% and 17.6\%, respectively, while the response cosine increases from $-0.73$ to $-0.31$. The corresponding statically merged policy produces substantially weaker redundant translation commands on BendPick, and the number of successful trials increases from 0/10 to 3/10.

These results indicate that constraining update responses helps mitigate post-merging capability degradation while preserving preliminary execution capabilities, establishing a more balanced starting point for subsequent training. Residual behavioral deviations still require further calibration. GeoMerge-OPD therefore builds on the merging of regularized experts and introduces OPD with task-specific teachers, providing supervision at the environment states and action-flow trajectories actually visited by the student to further integrate multitask skills.

\section{Method}
GeoMerge-OPD first trains task-specific experts independently with G-isometry regularization, then combines their parameters through static parameter merging to initialize a student with preliminary multitask execution capabilities. The corresponding task teachers subsequently provide velocity-field supervision at the environmental observations and action-generation states actually visited by the student, further correcting behavioral deviations left by the merge.

\noindent\begin{minipage}{\linewidth}
\centering
\includegraphics[width=\linewidth,trim={80bp 80bp 80bp 80bp},clip]{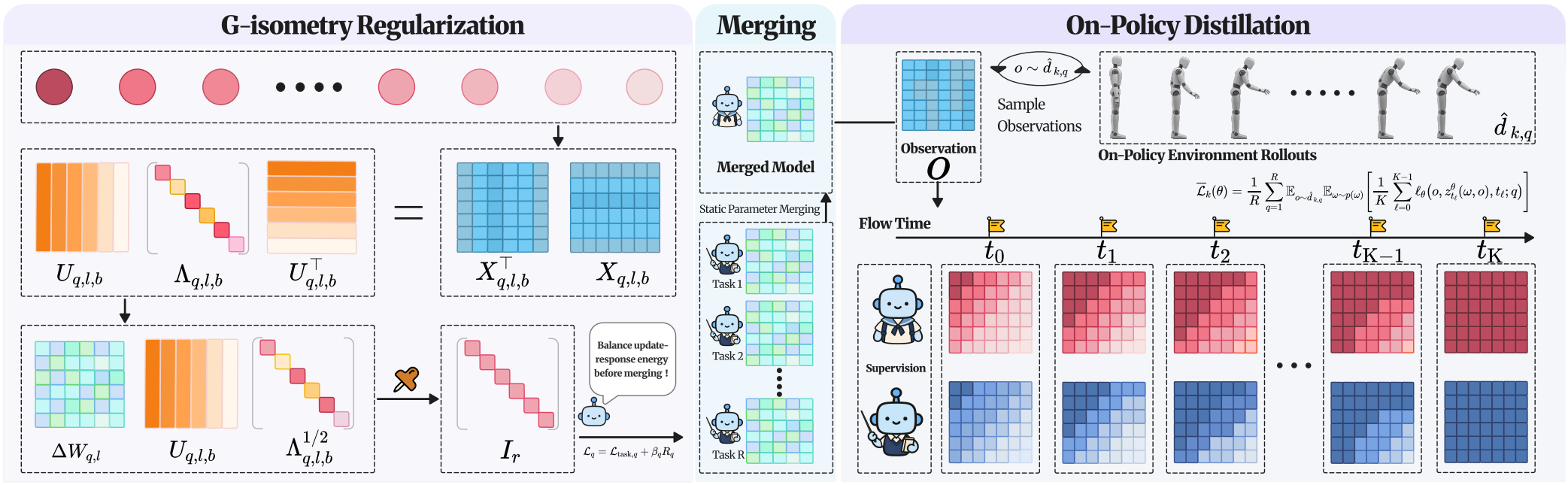}
\captionof{figure}{GeoMerge-OPD: G-isometry shapes task-specific update responses; static merging initializes one student; task teachers supervise its velocity field along student-visited observations and action-flow states.}
\label{fig:workflow}
\end{minipage}
\par

\subsection{G-Isometry Regularization}

The analysis in Section 3 shows that the functional effect of an expert update depends on the response it produces on task inputs. Accordingly, following the response-energy expression in Eq.~\eqref{eq:response-energy}, we constrain the response of each expert update in the principal activation subspace of its task.

Suppose there are $R$ task-specific experts, with $q$ indexing an expert and its corresponding task group. For input activations at layer $l$ and noise bin $b$, we combine the batch and token dimensions into a sample dimension, obtaining $X_{q,l,b} \in \mathbb{R}^{n_{q,l,b} \times d_{\mathrm{in}}}$. Here, $n_{q,l,b}$ is the number of activation samples used for the statistics, and $d_{\mathrm{in}}$ is the input dimension of the layer. We compute the Gram matrix normalized by the number of samples and retain its leading $r$ eigenvectors:

\begin{equation}
G_{q,l,b} = \frac{1}{n_{q,l,b}} X_{q,l,b}^{\top} X_{q,l,b} \approx U_{q,l,b} \Lambda_{q,l,b} U_{q,l,b}^{\top}.
\label{eq:gram-eigendecomposition}
\end{equation}

Here, $r$ is the number of retained directions, the columns of $U_{q,l,b}$ represent the principal activation directions, and $\Lambda_{q,l,b}$ is the diagonal matrix of their corresponding eigenvalues. Let $\Delta W_{q,l}$ denote the update at this layer relative to the pretrained weights, and define the weighted response

\begin{equation}
M_{q,l,b} = \Delta W_{q,l} U_{q,l,b} \Lambda_{q,l,b}^{1/2} \in \mathbb{R}^{d_{\mathrm{out}} \times r},
\label{eq:weighted-response-matrix}
\end{equation}

where $d_{\mathrm{out}}$ is the output dimension of the layer. Each column of $M_{q,l,b}$ represents the response of the weight update to a principal activation direction, weighted by the square root of that direction's activation energy so that the response scale reflects the role of different input directions in the task.

We penalize deviations of the weighted-response Gram matrix from the identity matrix:

\begin{equation}
R_{q,l,b} = \frac{1}{r} \left\lVert M_{q,l,b}^{\top} M_{q,l,b} - I_r \right\rVert_F^2.
\label{eq:g-isometry-penalty}
\end{equation}

Here, $I_r$ is the $r$-dimensional identity matrix, and $\lVert \cdot \rVert_F$ is the Frobenius norm. The diagonal entries of the response Gram matrix represent the energies of the column responses, while its off-diagonal entries represent their inner products. Bringing this matrix close to the identity constrains both response scale and directional correlation, mitigating excessive dominance of a few directions in the response energy.

Aggregating the penalties over all noise bins and constrained layers gives the task regularizer and training objective:

\begin{equation}
R_q = \sum_{b=1}^{m} w_b \left( \frac{1}{|\mathcal{A}|} \sum_{l \in \mathcal{A}} R_{q,l,b} \right), \qquad \mathcal{L}_q = \mathcal{L}_{\mathrm{task},q} + \beta_q R_q.
\label{eq:regularized-task-loss}
\end{equation}

Here, $m$ is the number of noise bins, and $\mathcal{A}$ is the set of constrained layers. $\mathcal{L}_{\mathrm{task},q}$ is the original task loss, and $\beta_q$ controls the regularization strength. The weights $w_b$ decrease monotonically as the noise level increases, imposing stronger constraints in lower-noise bins. All task-specific experts are fine-tuned independently from the same pretrained VLA backbone, yielding their respective parameters $\theta_q$.

\subsection{Parameter Merging and Student Initialization}

For task-specific experts derived from the same pretrained base, we merge both the vision-language and action-module parameters, including their LoRA updates, using a static merging operator $\mathcal{M}$, obtaining $\theta_{\mathrm{merge}} = \mathcal{M}(\{\theta_q\}_{q=1}^{R})$. The student is initialized with $\theta_{\mathrm{merge}}$, providing a starting point that retains preliminary multitask execution capabilities for subsequent distillation. Each task-specific expert retains its own parameters and serves as a fixed teacher during distillation, supervising the student as it corrects the remaining behavioral deviations after merging.

\subsection{On-Policy Distillation}

We first let the student interact with the environment and then generate actions under the observations it actually encounters, with the corresponding task teacher supervising the action-generation process. This involves two levels of sampling: the robot's execution trajectories in the environment and the internal denoising trajectories from noise to action during each action generation. The teacher provides a velocity field at the intermediate action states actually traversed by the student, helping it correct its current generation errors.

\subsubsection{Environment-Level Observation Distribution}

Let the student parameters be $\theta$, with initial parameters $\theta^{(0)} = \theta_{\mathrm{merge}}$. The student parameters before update round $k$ are $\theta^{(k)}$, and the corresponding action-generation distribution is $\pi_{\theta^{(k)}}$. Given an observation condition $o$, the student generates an action $u \sim \pi_{\theta^{(k)}}(\cdot \mid o)$ and interacts with the environment in closed loop.

For task group $q$, we collect $G$ episodes and save the observations at each action replanning step. Let episode $e$ contain $N_{k,q,e}$ such replanning observations. We construct the empirical distribution

\begin{equation}
\hat{d}_{k,q} = \frac{1}{G} \sum_{e=0}^{G-1} \frac{1}{N_{k,q,e}} \sum_{i=0}^{N_{k,q,e}-1} \delta_{o_{k,q,e,i}}.
\label{eq:empirical-observation-distribution}
\end{equation}

Here, $i$ indexes observations within an episode, and $\delta_o$ is a point mass at observation $o$. This distribution is normalized within each episode before averaging episodes with equal weight. It is then used to sample observation conditions for subsequent distillation.

\subsubsection{Action-Level Velocity-Field Distillation}

Given $o \sim \hat{d}_{k,q}$, we hold this observation fixed and rerun the student's action-generation process. In normalized action space, the student starts from standard Gaussian noise and progressively denoises it into an action. Using the flow time convention in which the noise endpoint is $t = 1$ and the action endpoint is $t = 0$, we divide denoising uniformly into $K$ steps, with $t_0 = 1 > \cdots > t_K = 0$. For $\ell = 0, \ldots, K - 1$, let $h_\ell = t_\ell - t_{\ell+1} > 0$. We denote by $z_{t_\ell}^{\theta}$ the intermediate latent action state actually traversed by the student at step $\ell$, while $z$ denotes a generic intermediate state.

At each step, we first compute the update mean from the student velocity field and then add a Gaussian perturbation:

\begin{equation}
z_{t_{\ell+1}}^{\theta} = \mu_{\theta}\!\left(z_{t_\ell}^{\theta}, o, t_\ell\right) + g_\ell \sqrt{h_\ell}\,\xi_\ell, \qquad \xi_\ell \sim \mathcal{N}(0, I).
\label{eq:student-sampling-step}
\end{equation}

Here, $g_\ell > 0$ is the diffusion coefficient, $\xi_\ell$ is an independent standard Gaussian perturbation at each step, and $I$ is the identity matrix with the same dimension as the latent action state. The student velocity field $v_\theta(z,t;o)$ predicts the flow velocity from the current action state, flow time, and observation, and is used to construct the conditional mean of the denoising update. The Euler-Maruyama discretization of the reverse SDE gives

\begin{equation}
\mu_\theta(z,o,t_\ell) = \left(1 - \frac{h_\ell g_\ell^2}{2t_\ell}\right) z - h_\ell \left(1 + \frac{g_\ell^2(1-t_\ell)}{2t_\ell}\right) v_\theta(z,t_\ell;o).
\label{eq:reverse-sde-mean}
\end{equation}

This update makes the next denoising state Gaussian conditional on the current state, observation, and time. To compare student and teacher denoising behavior, at the same $z,o,t_\ell$, we query the task teacher $v_{\theta_q}$ and use the same noise schedule and covariance. The teacher mean $\mu_{\theta_q}$ is obtained by substituting its velocity field into the equation above.

Let $C$ be the set of valid scalar action coordinates included in the loss, $P_C$ the corresponding coordinate-selection operator, and $D = |C|$. The student and teacher transition kernels on these coordinates can both be written as

\begin{equation}
K_\psi^C(\cdot \mid z,o,t_\ell) = \mathcal{N}\!\left( P_C \mu_\psi(z,o,t_\ell), g_\ell^2 h_\ell I_D \right), \qquad \psi \in \{\theta, \theta_q\},
\label{eq:projected-transition-kernels}
\end{equation}

where $\psi$ denotes the parameters of either the student or the current task teacher, and $I_D$ is the $D$-dimensional identity matrix.

Because the covariances are identical, the local KL divergence can be expressed as a difference between means and then reduced to a velocity-field error:

\begin{equation}
D_{\mathrm{KL}}\!\left(K_\theta^C \middle\| K_{\theta_q}^C\right) = \frac{\left\lVert P_C(\mu_\theta - \mu_{\theta_q}) \right\rVert_2^2} {2g_\ell^2 h_\ell} = w_\ell \left\lVert P_C(v_\theta - v_{\theta_q}) \right\rVert_2^2.
\label{eq:local-gaussian-kl}
\end{equation}

Here, $\lVert \cdot \rVert_2$ is the Euclidean norm. The abbreviated means and velocity fields are evaluated at the same observation, intermediate state, and flow time. The time-step weight $w_\ell=\frac{h_\ell}{2g_\ell^2}\left(1+\frac{g_\ell^2(1-t_\ell)}{2t_\ell}\right)^2$ converts velocity-field differences into distributional differences for the corresponding denoising step. Derivations of the sampling mean and KL divergence are provided in Appendix~\ref{app:opd}.

During distillation, the teacher provides supervision at the student-visited $z_{t_\ell}^{\theta}$. Normalizing by the number of valid action coordinates, we define the local loss

\begin{equation}
\ell_\theta\!\left(o,z_{t_\ell}^{\theta},t_\ell;q\right) = \frac{w_\ell}{D} \left\lVert P_C\!\left[ v_\theta\!\left(z_{t_\ell}^{\theta},t_\ell;o\right) - \operatorname{SG}\!\left[ v_{\theta_q}\!\left(z_{t_\ell}^{\theta},t_\ell;o\right) \right] \right] \right\rVert_2^2.
\label{eq:opd-point-loss}
\end{equation}

Here, $\operatorname{SG}$ treats the teacher output as a stop-gradient supervision target and also blocks the gradient of the teacher output with respect to its input state. Both the student velocity field and its action-generation path participate in backpropagation.

During the current update, the environment observation distribution $\hat{d}_{k,q}$ remains fixed. We collectively denote the initial noise for action generation and all stepwise perturbations by the base noise $\omega$, whose joint distribution $p(\omega)$ is independent of the parameters being optimized. The student path can therefore be reparameterized as a differentiable function $\mathbf z^\theta(\omega,o)$, giving the overall training objective

\begin{equation}
\bar{\mathcal{L}}_k(\theta) = \frac{1}{R} \sum_{q=1}^{R} \mathbb{E}_{o \sim \hat{d}_{k,q}} \mathbb{E}_{\omega \sim p(\omega)} \left[ \frac{1}{K} \sum_{\ell=0}^{K-1} \ell_\theta\!\left( o, z_{t_\ell}^{\theta}(\omega,o), t_\ell; q \right) \right].
\label{eq:expected-opd-objective}
\end{equation}

We compute the reparameterized gradient with the above stop-gradient operation, $\left.\nabla_\theta \bar{\mathcal{L}}_k(\theta)\right|_{\theta=\theta^{(k)}}$, update the student parameters, and alternate between environment sampling and distillation updates.

\setlength{\floatsep}{8pt plus 2pt minus 2pt}
\setlength{\textfloatsep}{11pt plus 2pt minus 4pt}
\setlength{\intextsep}{8pt plus 2pt minus 2pt}
\par\Needspace{4\baselineskip}
\section{Experiments}
\label{sec:experiments}
\begin{table}[!b]
\setlength{\aboverulesep}{1.2pt}
\setlength{\belowrulesep}{1.5pt}
\renewcommand{\MTgroup}[2]{\addlinespace[.3pt]\multicolumn{#1}{@{}l}{\textbf{#2}}\\[.1pt]}
\centering\MTsetup\renewcommand{\arraystretch}{1.0}\setlength{\abovecaptionskip}{1pt}
\caption{\textbf{Whole-body skill consolidation on SIMPLE ($\Psi_0$).} Success counts.}
\label{tab:main-whole-body}
\begin{tabularx}{\linewidth}{>{\raggedright\arraybackslash}Xc*{4}{c}}
\toprule
\textbf{Method} & \textbf{Init.} & \textbf{BendPick} & \textbf{CloseDoor} & \textbf{OpenOven} & \textbf{Total} \\
\midrule
\MTgroup{6}{Single-task Finetuned Model}
\rowcolor{MTgray}Standard FT (unreg.) & B & 9/10 & 10/10 & 10/10 & 29/30 \\
\rowcolor{MTgray}G-isometry FT & B & 10/10 & 10/10 & 10/10 & 30/30 \\
\midrule
\MTgroup{6}{Integration by static merging}
TIES & G & 2/10 & 4/10 & 1/10 & 7/30 \\
WUDI & G & 3/10 & 6/10 & 2/10 & 11/30 \\
RegMean++ & G & 3/10 & 5/10 & 4/10 & 12/30 \\
\midrule
\MTgroup{6}{Integration with post-training}
Joint FT & B & 0/10 & 6/10 & \textbf{8/10} & 14/30 \\
Joint FT & G & 5/10 & 9/10 & \textbf{8/10} & 22/30 \\
GRPO-mix & G & 5/10 & 7/10 & 7/10 & 19/30 \\
\midrule
\rowcolor{MTblue}\textbf{GeoMerge-OPD (Ours)} & G & \textbf{8/10} & \textbf{10/10} & \textbf{8/10} & \textbf{26/30} \\
\bottomrule
\end{tabularx}
\MTnote{B: pretrained base. S: unregularized experts. G: G-isometry experts (static merging) or their RegMean++ merge (post-training). Joint FT (G) and GRPO-mix share our G initialization. Joint FT and OPD both fully fine-tune the action backbone and interfaces. GRPO-mix and GeoMerge-OPD match preset episodes, updates, and observation exposures. OPD additionally uses teacher velocity-field supervision. Bold marks the best integration results, including ties.}
\end{table}

\subsection{Experimental setup}
\noindent\textbf{Benchmarks and evaluation.} We evaluate $\Psi_0$ on SIMPLE \citep{simple} for whole-body control and $\pi_{0.5}$ on LIBERO \citep{libero} and LIBERO-PRO \citep{liberopro} for manipulation. Task-specific SIMPLE and suite-specific LIBERO models are independently fine-tuned from their respective pretrained bases. Evaluation uses ten episodes per SIMPLE task (BendPick, CloseDoor, OpenOven) and 100 per LIBERO suite per evaluation seed over three evaluation seeds. LIBERO-PRO tests these policies under perturbations without retraining. We report equally weighted task, suite, or perturbation averages.

\noindent\textbf{Comparisons and training.} Task- and suite-specific fine-tuned models serve as references. TIES, WUDI, and RegMean++ use official implementations adapted only to $\Psi_0$ and $\pi_{0.5}$ and are evaluated after merging. Joint fine-tuning (Joint FT) is trained to convergence. GRPO-mix \citep{grpo} and GeoMerge-OPD match preset episodes, optimizer updates, and observation exposures. OPD additionally uses teacher velocity-field supervision. Expert and joint fine-tuning use VLM LoRA with $r=\alpha=64$ and fully fine-tune learnable interfaces. Both use a global batch size of 128, with learning rates of $3\times10^{-4}$ for $\pi_{0.5}$ and $1\times10^{-4}$ for $\Psi_0$, and full configurations in Appendix~\ref{app:experimental-setup}.

\begin{table}[!t]
\setlength{\aboverulesep}{1.2pt}
\setlength{\belowrulesep}{1.5pt}
\renewcommand{\MTgroup}[2]{\addlinespace[.3pt]\multicolumn{#1}{@{}l}{\textbf{#2}}\\[.1pt]}
\centering\MTsetup\renewcommand{\arraystretch}{1.0}\setlength{\abovecaptionskip}{1pt}
\caption{\textbf{Consolidation and robustness with $\pi_{0.5}$.} Mean success rates (\%) over three evaluations.}
\label{tab:transfer-pro}
\begingroup\fontsize{9}{10.4}\selectfont\setlength{\tabcolsep}{2pt}\renewcommand{\arraystretch}{1.0}
\begin{tabularx}{\linewidth}{>{\raggedright\arraybackslash}Xc*{5}{c}@{\hspace{5pt}}*{6}{c}}
\toprule
 & & \multicolumn{5}{c}{\textbf{LIBERO}} & \multicolumn{6}{c}{\textbf{LIBERO-PRO}} \\\cmidrule(lr){3-7}\cmidrule(lr){8-13}
\textbf{Method} & \textbf{Init.} & \textbf{Spat.} & \textbf{Obj.} & \textbf{Goal} & \textbf{Long} & \textbf{Avg.} & \textbf{Obj.} & \textbf{Pos.} & \textbf{Sem.} & \textbf{Task} & \textbf{Env.} & \textbf{Avg.} \\
\midrule
\MTgroup{13}{Single-task Finetuned Model}
\rowcolor{MTgray}Standard FT (unreg.) & B & 99.3 & 100.0 & 99.0 & 99.7 & 99.5 & 87.3 & 23.0 & 96.0 & 5.7 & 53.7 & 53.1 \\
\rowcolor{MTgray}G-isometry FT & B & 100.0 & 98.7 & 99.0 & 99.3 & 99.3 & 86.7 & 22.0 & 97.0 & 7.7 & 55.0 & 53.7 \\
\midrule
\MTgroup{13}{Integration by static merging}
TIES & G & 77.3 & 74.7 & 66.3 & 50.7 & 67.3 & 42.3 & 17.3 & 57.0 & 14.3 & 18.3 & 29.9 \\
WUDI & G & 65.3 & 61.3 & 65.0 & 43.7 & 58.8 & 38.3 & 10.3 & 45.7 & 20.0 & 13.0 & 25.5 \\
RegMean++ & G & 86.3 & 84.7 & 66.0 & 55.7 & 73.2 & 58.0 & 25.0 & 55.7 & 16.0 & 29.7 & 36.9 \\
\midrule
\MTgroup{13}{Integration with post-training}
Joint FT & B & \textbf{97.3} & 95.7 & \textbf{97.0} & 94.7 & 96.2 & 85.3 & 33.3 & 97.0 & 42.0 & 64.7 & 64.5 \\
Joint FT & G & 97.0 & 95.7 & 96.3 & 91.7 & 95.2 & \textbf{86.0} & 35.3 & \textbf{97.7} & 40.7 & 65.0 & 64.9 \\
GRPO-mix & G & 93.3 & 91.7 & 90.7 & 84.7 & 90.1 & 79.3 & 37.7 & 89.0 & 33.7 & 59.3 & 59.8 \\
\midrule
\rowcolor{MTblue}\textbf{GeoMerge-OPD} & G & \textbf{97.3} & \textbf{98.7} & 95.3 & \textbf{97.0} & \textbf{97.1} & 85.3 & \textbf{42.7} & 95.0 & \textbf{46.7} & \textbf{67.0} & \textbf{67.3} \\
\bottomrule
\end{tabularx}
\endgroup
\MTnote{Codes and boldface follow Table~\ref{tab:main-whole-body}. Avg. equally weights four LIBERO suites or five PRO axes. Sem.: language paraphrasing.}
\end{table}

\subsection{Whole-body skill integration}
All three static merging operators retain preliminary execution capabilities on every SIMPLE task after merging G-isometry-regularized models (Table~\ref{tab:main-whole-body}). RegMean++ yields the highest static average and most balanced task profile, so its merged model is our default post-training initialization. We also initialize Joint FT from this G-regularized merge to construct a strengthened baseline sharing our starting point. This raises Joint FT from 14/30 with the pretrained base to 22/30, extending the benefit of our initialization to joint fine-tuning. From the same initialization, GeoMerge-OPD reaches 26/30 and is best or tied-best across all three tasks among the evaluated integration methods, yielding strong, balanced whole-body performance.

\Needspace{22\baselineskip}
\subsection{Cross-backbone evaluation and robustness}
\begin{wrapfigure}{r}{0.42\textwidth}
\vspace{-9pt}
\centering
\includegraphics[width=\linewidth,trim={0bp 0bp 0bp 45bp},clip]{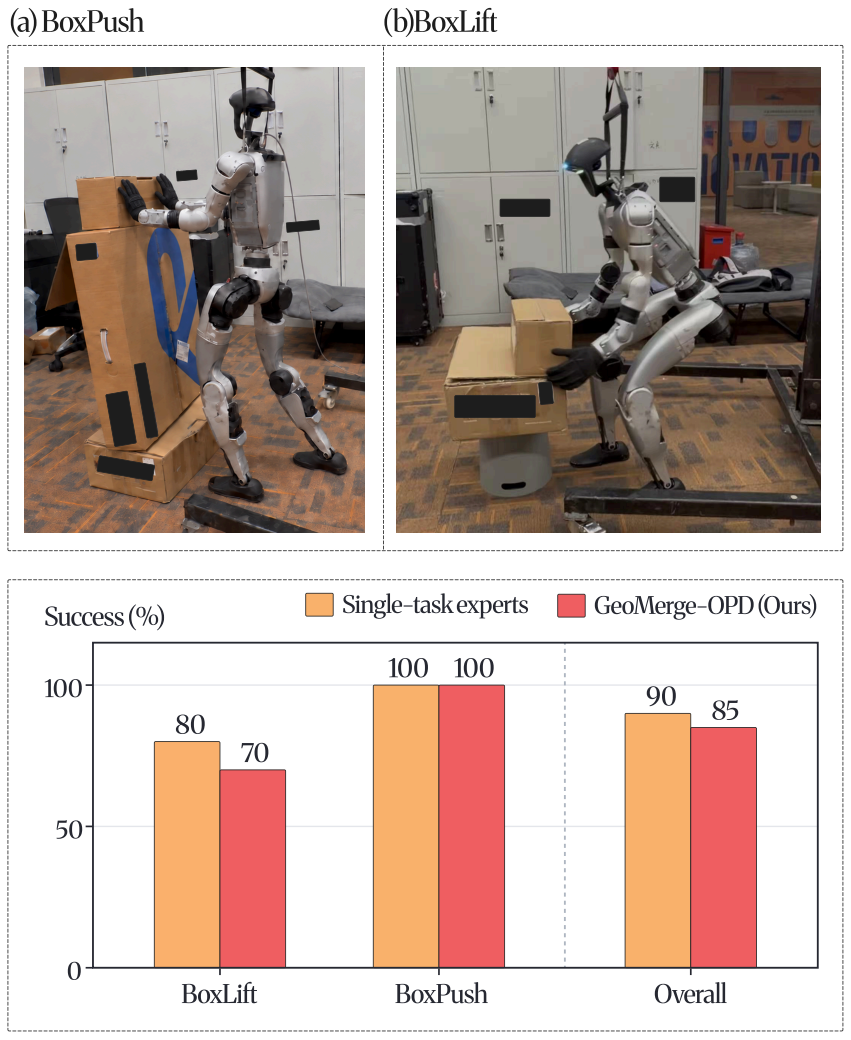}
\par\vspace{-9pt}
\setlength{\abovecaptionskip}{4pt}
\caption{\small Unitree G1 evaluation, with 10 trials per policy per task.}
\label{fig:real-world}
\end{wrapfigure}
We extend evaluation from $\Psi_0$ whole-body control to $\pi_{0.5}$ manipulation (Table~\ref{tab:transfer-pro}). GeoMerge-OPD achieves the highest average among evaluated multi-skill integration methods on LIBERO and LIBERO-PRO, at 97.1\% and 67.3\%. On LIBERO, its average surpasses static merging, GRPO-mix, and joint fine-tuning from the pretrained base. On LIBERO-PRO, its average exceeds the 64.9\% of joint fine-tuning from the G-regularized merge, with the best scores on position perturbations, task variations, and environment changes. That baseline remains slightly higher on object perturbations and language paraphrasing, by 0.7 and 2.7 percentage points, respectively. These results support benefits across VLA backbones and strong shared-policy performance under changing task conditions.

\subsection{Real-world experiments}

\noindent\textbf{Tasks.} We evaluate GeoMerge-OPD on a Unitree G1 humanoid robot in two stationary manipulation tasks (Fig.~\ref{fig:real-world}). BoxPush requires pushing an elevated cardboard box forward a short distance, and BoxLift requires bending down to lift a box from a low platform and hold it stably. We collect 50 teleoperated demonstrations per task for fine-tuning. Each policy is evaluated over 10 trials per task using the same observation interface and low-level controller. Success requires autonomous task completion.

\noindent\textbf{Results.} GeoMerge-OPD achieves 85\% overall success (17/20), compared with 90\% (18/20) for the single-task experts. Both achieve 100\% on BoxPush. On BoxLift, GeoMerge-OPD achieves 70\%, compared with 80\% for the corresponding specialist. These results provide preliminary evidence that one shared action expert can retain both real-world skills.

\par

\par
\ifnum\value{WF@wrappedlines}>1
  \vspace{\dimexpr\value{WF@wrappedlines}\baselineskip-\baselineskip\relax}
\fi
\WFclear

\Needspace{5\baselineskip}
\subsection{Ablations}
\begin{figure}[htbp]
\centering
\setlength{\abovecaptionskip}{5pt}
\includegraphics[width=\linewidth,trim={60bp 60bp 60bp 60bp},clip]{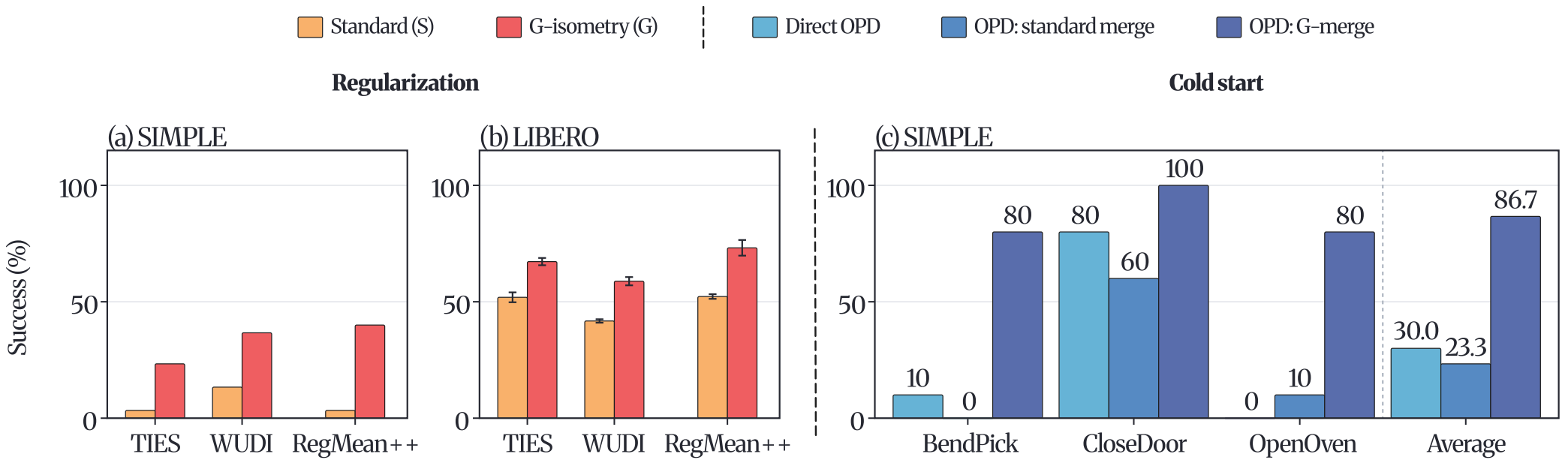}
\caption{(a,b) Static merging from standard fine-tuned models (S, unregularized) and G-isometry fine-tuned models (G). LIBERO error bars show sample SD of seed-level suite averages. (c) SIMPLE success after direct OPD or OPD from standard (unregularized) and G-isometry merged initializations. The teachers are unchanged across these initializations. Average equally weights the three tasks.}
\label{fig:ablations}
\end{figure}

\noindent\textbf{Regularization before merging.} G-isometry improves the average static-merging success of TIES, WUDI, and RegMean++ on both SIMPLE and LIBERO (Figure~\ref{fig:ablations}a,b). This consistency across operators supports adjusting task-specific parameter updates before merging. The resulting task competence provides the foundation for subsequent behavior-level refinement.

\noindent\textbf{Initialization for on-policy distillation.} OPD from the G-regularized merge achieves 86.7\% mean success, compared with 30.0\% for direct OPD and 23.3\% for OPD from a standard (unregularized) merge (Figure~\ref{fig:ablations}c). The G-initialized policy retains all three skills, whereas each alternative exhibits complete collapse on one of the evaluated tasks. Appendix~\ref{app:ablation-results} adds regularization and distillation comparisons, observation-source and regularization-strength ablations, and results across three post-training seeds. Appendix~\ref{app:mechanism-diagnostics} extends the response-geometry diagnostic to OrthoReg.

\section{Conclusion}
GeoMerge-OPD combines response-space regularization, parameter merging, and on-policy distillation into a complete framework for integrating heterogeneous robotic skills. G-isometry improves the mergeability of expert updates during independent fine-tuning, while OPD uses task teachers to calibrate post-merging behavioral deviations on observations and action-generation trajectories actually visited by the student. Experiments with two VLA backbones demonstrate balanced multi-skill performance, and LIBERO-PRO further shows robustness under perturbations. Real-world experiments further demonstrate the feasibility of integrating multiple manipulation skills into a shared policy on a humanoid robot.

\textbf{Limitations and future work.} Current real-world validation mainly covers two stationary manipulation tasks, and the expert combinations studied are limited in scale. Future work will extend to real-world tasks that alternate locomotion and manipulation and involve richer contact modes, and examine skill retention, balance, and robustness as the number of experts increases. We will also conduct more extensive evaluations across diverse scenarios and independent runs.

\label{maintext-end}
\clearpage
\section*{AI use statement}
Generative AI tools (Codex) assisted with translation, manuscript editing, consistency checks, and summarizing and visualizing author-provided experimental results. The authors take full responsibility for the final manuscript.

\bibliography{references}
\bibliographystyle{iclr2027_conference}
\clearpage
\appendix
\raggedbottom
\numberwithin{equation}{section}
\section{Training and Method Settings}
\label{app:training-method-settings}
\label{app:experimental-setup}

\label{app:a:training-settings}

We use PaliGemma with a Gemma action expert for $\pi_{0.5}$ and
Qwen3-VL-2B with an action expert for $\Psi_0$. All training runs use proprioception and
four A100 80GB GPUs. The original experiments use one training seed.
Appendix~\ref{app:training-seeds} additionally reports three post-training
seeds for GeoMerge-OPD and Joint FT. For G-initialized methods, the experts
and merged initialization are held fixed across these runs. These training
seeds are distinct from the repeated evaluation seeds used for LIBERO
and LIBERO-PRO.

\par\smallskip
\begingroup
\fontsize{9}{10.4}\selectfont
\setlength{\tabcolsep}{3pt}
\renewcommand{\arraystretch}{1.0}
\begin{minipage}{\linewidth}
\captionof{table}{Expert and joint fine-tuning settings.}
\label{tab:appendix-expert-settings}
\noindent\begin{tabularx}{\linewidth}{@{}>{\raggedright\arraybackslash}p{0.29\linewidth}*{4}{>{\centering\arraybackslash}X}@{}}
\toprule
\rowcolor{MTblue}\textbf{Setting} & \shortstack{$\pi_{0.5}$\\\textbf{Expert}} & \shortstack{$\pi_{0.5}$\\\textbf{Joint FT}} & \shortstack{$\Psi_0$\\\textbf{Expert}} & \shortstack{$\Psi_0$\\\textbf{Joint FT}}\\
\midrule
Training scope & One suite & Four suites & One task & Three tasks\\
Effective global batch & 128 & 128 & 128 & 128\\
Learning rate & $3\!\times\!10^{-4}$ & $3\!\times\!10^{-4}$ & $1\!\times\!10^{-4}$ & $1\!\times\!10^{-4}$\\
VLM tuning & LoRA & LoRA & LoRA & LoRA\\
LoRA rank $r$ & 64 & 64 & 64 & 64\\
LoRA scale $\alpha$ & 64 & 64 & 64 & 64\\
Action backbone & LoRA & Full & LoRA & Full\\
Interfaces & Full & Full & Full & Full\\
Training steps & 15,000 & 30,000 & 30,000 & 60,000\\
Effective images & 2 & 2 & 1 & 1\\
\bottomrule
\end{tabularx}%
\end{minipage}
\endgroup\par\smallskip
\noindent Image counts exclude empty cameras. Joint FT samples the three
$\Psi_0$ tasks equally.

\label{app:a:method-settings}

TIES, WUDI, and RegMean++ follow official implementations
\citep{ties,wudi,regmeanpp}, adapted only to the two backbones.
OPD and GRPO-mix share the G-regularized merge and matched episode, update,
and observation-exposure budgets. OPD adds teacher supervision.

\par\medskip
\begingroup
\fontsize{9}{10.4}\selectfont
\setlength{\tabcolsep}{3pt}
\renewcommand{\arraystretch}{1.0}
\begin{minipage}{\linewidth}
\captionof{table}{Regularization, merging, and post-training settings.}
\label{tab:appendix-integration-settings}
\noindent\begin{tabularx}{\linewidth}{@{}>{\raggedright\arraybackslash}p{0.40\linewidth}*{2}{>{\centering\arraybackslash}X}@{}}
\toprule
\rowcolor{MTblue}\textbf{Setting} & $\pi_{0.5}$ & $\Psi_0$\textbf{ (three skills)}\\
\midrule
\rowcolor{MTblue!45}\textbf{G-isometry} & & \\
Update / $r_G$ / equal-width bins & $\Delta W=BA$ / 64 / 8 & $\Delta W=BA$ / 64 / 8\\
Training obs. per expert / Gram & 1,024 / fixed & 1,024 / fixed\\
Regularization strength $\beta_q$ & 8 & 2\\
\rowcolor{MTblue!45}\textbf{Merging} & & \\
Operator / $\alpha_{\mathrm{RM}}$ / ridge & RegMean++ / 0.3 / 0 & RegMean++ / 0.3 / 0\\
Experts / weight per expert & 4 / $\tfrac14$ & 3 / $\tfrac13$\\
Calibration obs. per expert / total & 1,024 / 4,096 & 1,024 / 3,072\\
Gram updates & Blockwise & Blockwise\\
\rowcolor{MTblue!45}\textbf{OPD / GRPO-mix} & & \\
VLM LoRA / interfaces / action backbone & $r=\alpha=64$ / Full / Full & $r=\alpha=64$ / Full / Full\\
Module/interface LR: VLM / action & $10^{-6}$ / $10^{-5}$ & $10^{-6}$ / $10^{-5}$\\
Optimizer / observation batch & AdamW / 128 & AdamW / 128\\
Episodes per group / updates per round & 2 / 1 (current round only) & 2 / 1 (current round only)\\
Optimizer updates / collected episodes & 800 / 6,400 & 1,000 / 6,000\\
Observation exposures & 102,400 & 128,000\\
Action chunk length / executed steps & 50 / 10 & 30 / 24\\
OPD loss / GRPO stochastic dims. & 350 / 1,600 & \shortstack{1,080 / 1,080 (initial)\\864 / 864 (6-step RTC prefix)}\\
Uniform flow steps $K$ / noise scale $\eta$ & 10 / 0.1 & 10 / 0.1\\
\rowcolor{MTblue!45}\textbf{GRPO} & & \\
Group size / standardized-advantage $\epsilon$ & 2 / $10^{-8}$ & 2 / $10^{-8}$\\
Clipping / reference-KL coefficient & 0.2 / 0.01 & 0.2 / 0.01\\
Reward / reference policy & Binary success / initial merge & Binary success / initial merge\\
\bottomrule
\end{tabularx}%
\end{minipage}
\endgroup\par\smallskip

G-isometry uses Gram matrices precomputed from the common pretrained base.
GRPO groups independent trajectories sharing a task and complete initial
state. Observation exposures count draws with replacement. The unnormalized
bin weights are $w_b=\exp[-(b-0.5)/(8\times0.3)]/8$, $b=1,\ldots,8$.
Sampling and reductions follow Appendix~\ref{app:algorithm-details}.
With $t_\ell=1-\ell/K$ and $h_\ell=0.1$, the diffusion coefficient is
\begin{equation}
g_\ell=\eta\sqrt{\frac{t_\ell}{1-\bar t_\ell}},
\qquad
\bar t_\ell=
\begin{cases}
t_{\ell+1}, & t_\ell=1,\\
t_\ell, & t_\ell<1,
\end{cases}
\quad \ell=0,\ldots,K-1.
\label{app:eq:noise-schedule}
\end{equation}
Only the denominator substitutes $t_1$ at $t_0=1$. Drift and local-KL
weights retain $t_\ell$.

\FloatBarrier
\Needspace{10\baselineskip}
\renewcommand{\thealgorithm}{B.\arabic{algorithm}}
\section{Algorithm and Method Details}
\label{app:algorithm-details}

This section details the flow of parameters, observations, and gradients
through expert training, merging, and two-level on-policy distillation.
We use $R$ task groups, indexed by $q$, update round $k$, and flow step
$\ell$. The training settings are specified separately. The algorithm below
uses the objectives and sampling convention of the main text.

\subsection{Expert Training and Student Initialization}
\label{app:b:initialization}

Each task expert is independently fine-tuned from the same pretrained
base. Task activation statistics define its principal input directions and
their energies. G-isometry acts on the corresponding weighted response of
the expert update relative to that base. The expert for task group $q$
minimizes $\mathcal L_{\mathrm{task},q}+\beta_qR_q$, where $R_q$ aggregates
the response penalties over constrained layers and noise bins.

The static operator combines the experts' vision-language and action-module
parameters, including their LoRA updates, to form
$\theta_{\mathrm{merge}}=\mathcal M(\{\theta_q\}_{q=1}^{R})$.
The shared student is initialized with
$\theta^{(0)}=\theta_{\mathrm{merge}}$.
The experts retain their own parameters and serve as fixed task teachers.
The student provides a single policy for execution and receives subsequent
updates. For observations from task group $q$, teacher $\theta_q$ provides
the velocity-field targets. The task index organizes teacher supervision,
while the resulting actions are generated by the shared student.

\begin{center}
\begin{tabular}{@{}c@{\hspace{7pt}}c@{\hspace{7pt}}c@{\hspace{7pt}}c@{\hspace{7pt}}c@{}}
\textbf{Train experts} & $\longrightarrow$ & \textbf{Merge parameters}
& $\longrightarrow$ & \textbf{Distill the student}\\[-1pt]
G-isometry && VLA initialization && two-level on-policy sampling
\end{tabular}
\end{center}

\subsection{Two-Level Sampling and Episode Weights}
\label{app:b:sampling}

\paragraph{Environment-level observations.}
For $\pi_{0.5}$, a task group is one LIBERO suite, with its ten tasks selected in cyclic order across rounds. For $\Psi_0$, each group is one SIMPLE task.
Before update round $k$, student $\theta^{(k)}$ interacts with the
environment in closed loop using $\pi_{\theta^{(k)}}$. For each task group,
it collects $G$ episodes and saves the observation at each action
replanning step. These observations reflect the environment states reached
through the student's preceding actions. If episode $e$ contains
$N_{k,q,e}$ saved observations, the empirical distribution is
\begin{equation}
\hat d_{k,q}
=\frac1G\sum_{e=0}^{G-1}\frac1{N_{k,q,e}}
 \sum_{i=0}^{N_{k,q,e}-1}\delta_{o_{k,q,e,i}}.
\label{app:b:empirical-observations}
\end{equation}
Here $\delta_o$ is a point mass at $o$. Sampling from this distribution
means selecting an episode uniformly, then selecting one of its saved
replanning observations uniformly. Each episode therefore receives equal
total weight. Uniform sampling from a concatenated list of all observations
would instead give longer episodes more weight and define a different
training distribution.

\paragraph{Action-level resampling.}
For a sampled observation $o\sim\hat d_{k,q}$, we hold $o$ fixed and rerun
the student's action-generation process with fresh Gaussian noise.
At every intermediate state $z_{t_\ell}^{\theta}$, the student and the
corresponding teacher receive the same observation, intermediate state,
and flow time. The teacher supplies a velocity target at a state generated
by the student.

\begin{center}
\begin{minipage}{\linewidth}
\captionof{table}{Environment execution and flow resampling.}
\label{tab:appendix-sampling-levels}
\renewcommand{\arraystretch}{1.15}
\begin{tabular}{@{}>{\raggedright\arraybackslash}p{0.47\linewidth}@{\hspace{0.04\linewidth}}>{\raggedright\arraybackslash}p{0.47\linewidth}@{}}
\toprule
\textbf{Environment execution} & \textbf{Flow resampling for supervision}\\
\midrule
Execute the student's actions and obtain subsequent observations.
& Fix one saved observation and draw a new action-generation path.\\
Save replanning observations with their episode identities.
& Query the teacher at each student-visited intermediate action state.\\
Define the empirical observation distribution $\hat d_{k,q}$.
& Compute the loss along a differentiable student flow path.\\
\bottomrule
\end{tabular}
\end{minipage}
\end{center}

The action-generation paths used during environment execution determine
which observations are reached. The paths resampled at those observations
provide the distillation loss. Their connection is the saved observation
condition. Resampling does not require reusing the execution-time noise
or reproducing the execution-time intermediate states.

In normalized action space, $z_{t_0}\sim\mathcal N(0,I)$ and the flow times
are uniform with $t_0=1>\cdots>t_K=0$. Let
$h_\ell=t_\ell-t_{\ell+1}>0$. The resampled student path follows
\begin{equation}
z_{t_{\ell+1}}^\theta
=\mu_\theta(z_{t_\ell}^{\theta},o,t_\ell)
 +g_\ell\sqrt{h_\ell}\,\xi_\ell,
\qquad \xi_\ell\sim\mathcal N(0,I),
\label{app:b:student-step}
\end{equation}
where $g_\ell>0$ and the perturbations are independent. The reverse-SDE
mean used in the main text is
\begin{equation}
\begin{aligned}
\mu_\theta(z,o,t_\ell)
={}&\left(1-\frac{h_\ell g_\ell^2}{2t_\ell}\right)z\\
&-h_\ell\left(1+\frac{g_\ell^2(1-t_\ell)}{2t_\ell}\right)
  v_\theta(z,t_\ell;o).
\end{aligned}
\label{app:b:sampling-mean}
\end{equation}

\subsection{Loss Reduction and Gradient Flow}
\label{app:b:loss-gradients}

The student and teacher use the same noise schedule and covariance when
comparing their transitions at a common $o,z,t_\ell$. The equal-covariance
Gaussian KL is therefore a time-weighted squared velocity difference.
Let $C$ contain the valid scalar action coordinates, let $P_C$ select those
coordinates, and let $D=|C|$. Subsequent $\Psi_0$ chunks hold the six-step
RTC prefix fixed and exclude it from the loss and stochastic kernel. The local loss is
\begin{equation}
\begin{aligned}
\ell_\theta(o,z_{t_\ell}^{\theta},t_\ell;q)
={}&\frac{w_\ell}{D}\Big\|P_C\Big[
 v_\theta(z_{t_\ell}^{\theta},t_\ell;o)\\
&\hspace{33pt}
 -\operatorname{SG}\big[v_{\theta_q}(z_{t_\ell}^{\theta},t_\ell;o)\big]
 \Big]\Big\|_2^2,
\end{aligned}
\label{app:b:point-loss}
\end{equation}
with the main-text time-step weight
\[
w_\ell=\frac{h_\ell}{2g_\ell^2}
\left(1+\frac{g_\ell^2(1-t_\ell)}{2t_\ell}\right)^2.
\]
The factor $1/D$ averages the sum of squared errors over valid coordinates.
The full objective additionally averages over the $K$ flow steps and the
$R$ task groups. These are separate reductions: once a coordinate mean has
been computed, the same coordinate count is not divided out again.

During the current update, $\hat d_{k,q}$ is fixed. Let $\omega$ collect the
initial noise and all stepwise perturbations, with a distribution
$p(\omega)$ independent of the optimized parameters. The resampled student
path is a differentiable function $\mathbf z^\theta(\omega,o)$. The
reparameterized objective is
\begin{equation}
\begin{aligned}
\bar{\mathcal L}_k(\theta)
={}&\frac1R\sum_{q=1}^{R}
\mathbb E_{o\sim\hat d_{k,q}}\mathbb E_{\omega\sim p(\omega)}\\
&\quad\left[\frac1K\sum_{\ell=0}^{K-1}
\ell_\theta\big(o,z_{t_\ell}^{\theta}(\omega,o),t_\ell;q\big)\right].
\end{aligned}
\label{app:b:expected-objective}
\end{equation}
Observation and noise samples estimate the two expectations within each
task group. The $\Psi_0$ batch rotates task quotas of 43/43/42. Losses are
averaged within each episode, then equally across episodes and task groups.

Gradients reach the student parameters through both the velocity prediction
at the current intermediate state and the earlier student flow updates
that produced this state. Stop-gradient is applied to the complete teacher
output. It blocks derivatives with respect to teacher parameters and the
teacher output's derivative with respect to its input state. The student
field and student path retain their computation graph. Environment
interaction and the saved observations are fixed conditions for the
current update, and no derivative is propagated through the past
environment execution history.

Evaluating the resampled path at the pre-update parameter values specifies
where the gradient is computed. This evaluation does not detach the student path.
With the teacher-output stop-gradient in Eq.~\eqref{app:b:point-loss}, the
algorithm evaluates
$\left.\nabla_\theta\bar{\mathcal L}_k(\theta)
\right|_{\theta=\theta^{(k)}}$, updates the student once, and then returns to
environment sampling for the next round. Appendix~\ref{app:opd} derives the
local Gaussian KL and its relationship to path-level errors and task
returns. The reduction and gradient rules above specify the training
objective used here.

\subsection{Complete Algorithm}
\label{app:b:complete-algorithm}

Algorithm~\ref{app:b:algorithm} connects the three stages. Each OPD round
collects environment observations, resamples student flow paths at those
observations, aggregates the loss, and performs one student update. The
per-task teachers remain fixed throughout distillation.

\begingroup
\setlength{\intextsep}{0pt}
\setlength{\topsep}{0pt}
\setlength{\partopsep}{0pt}
\begin{algorithm}[H]
\caption{GeoMerge-OPD training and integration}
\label{app:b:algorithm}
\begin{algorithmic}[1]
\Require Common pretrained base $\theta_0$, task data $\{\mathcal D_q\}_{q=1}^{R}$,
         merge operator $\mathcal M$, and training settings
\Ensure A shared student $\theta$
\Statex \textbf{Stage 1: independent regularized experts}
\For{$q=1,\ldots,R$}
  \State Initialize expert $\theta_q$ from the common base
  \State Construct G-isometry from the task activation statistics
  \State Independently train with $\mathcal L_{\mathrm{task},q}+\beta_qR_q$
\EndFor
\Statex \textbf{Stage 2: merging and student initialization}
\State Merge vision-language and action-module parameters, including LoRA updates:
       $\theta_{\mathrm{merge}}\gets\mathcal M(\{\theta_q\}_{q=1}^{R})$
\State Initialize student $\theta\gets\theta_{\mathrm{merge}}$
\State Preserve and fix the task teachers $\{\theta_q\}_{q=1}^{R}$
\Statex \textbf{Stage 3: two-level on-policy distillation}
\For{each update round $k$}
  \State Set $\theta^{(k)}\gets\theta$ and collect $G$ environment episodes per task group
  \State Save replanning observations and construct $\hat d_{k,q}$ by
         Eq.~\eqref{app:b:empirical-observations}
  \For{each task group $q$ and each of its observation samples}
    \State Uniformly sample an episode, then a saved observation $o$ within it
    \State Hold $o$ fixed and draw fresh $\omega=(z_{t_0},\xi_0,\ldots,\xi_{K-1})$
    \State Initialize the path with the sampled $z_{t_0}$ and retain its computation graph
    \For{$\ell=0,\ldots,K-1$}
      \State $v_\ell\gets v_\theta(z_{t_\ell}^{\theta},t_\ell;o)$
      \State $y_\ell\gets\operatorname{SG}[v_{\theta_q}(z_{t_\ell}^{\theta},t_\ell;o)]$
      \State Accumulate $\frac{w_\ell}{D}
             \lVert P_C(v_\ell-y_\ell)\rVert_2^2$
      \State Advance the differentiable state using Eq.~\eqref{app:b:student-step}
    \EndFor
  \EndFor
  \State Average observation/noise samples and flow steps within each task
  \State Average task-group losses as in Eq.~\eqref{app:b:expected-objective}
  \State Backpropagate through the student field and path at $\theta=\theta^{(k)}$
  \State Perform one update of the student parameters
\EndFor
\State \Return $\theta$
\end{algorithmic}
\end{algorithm}
\endgroup

\FloatBarrier
\Needspace{10\baselineskip}
\section{Detailed Ablation and Reliability Results}
\label{app:ablation-results}

We organize the additional evidence around three questions: how regularization
affects expert mergeability (Appendix~\ref{app:regularization-experiments}),
how the distillation design affects post-merging refinement
(Appendix~\ref{app:distillation-design}), and how final gains vary across
training runs (Appendix~\ref{app:training-seeds}). The original task- and
suite-level results underlying Figure~\ref{fig:ablations} are included with
the corresponding comparisons. S denotes standard fine-tuning without
G-isometry, and G denotes G-isometry-regularized fine-tuning.

\subsection{Regularization and Expert Mergeability}
\label{app:regularization-experiments}

We first examine how expert training affects static merging, then compare
regularization methods and vary the regularization strength.

\subsubsection{Effects Across Merging Operators}
Tables~\ref{tab:ablation-static-simple} and~\ref{tab:ablation-static-libero} evaluate static merging before OPD. Source identifies the fine-tuned input models. LIBERO uses 100 episodes per suite per seed. Its Avg. equally weights the four suites, and the associated SD is computed across their three evaluation-seed means, assuming corresponding seed order across suites.

\begin{table}[H]
\centering
\fontsize{9}{10.4}\selectfont\setlength{\tabcolsep}{3pt}\renewcommand{\arraystretch}{1.06}
\caption{Static merging on SIMPLE ($\Psi_0$), underlying Figure~\ref{fig:ablations}a. Task entries are successful episodes out of ten. Avg. is the task-macro success rate (\%).}
\label{tab:ablation-static-simple}
\begin{tabularx}{\linewidth}{@{}>{\raggedright\arraybackslash}Xc*{4}{c}@{}}
\toprule
\textbf{Merger} & \textbf{Source} & \textbf{BendPick} & \textbf{CloseDoor} & \textbf{OpenOven} & \textbf{Avg.} \\
\midrule
TIES & S & 0/10 & 1/10 & 0/10 & 3.3 \\
WUDI & S & 0/10 & 2/10 & 2/10 & 13.3 \\
RegMean++ & S & 0/10 & 0/10 & 1/10 & 3.3 \\
\midrule
TIES & G & 2/10 & 4/10 & 1/10 & 23.3 \\
WUDI & G & 3/10 & 6/10 & 2/10 & 36.7 \\
RegMean++ & G & 3/10 & 5/10 & 4/10 & 40.0 \\
\bottomrule
\end{tabularx}
\end{table}

\begin{table}[H]
\centering
\fontsize{9}{10.4}\selectfont\setlength{\tabcolsep}{3pt}\renewcommand{\arraystretch}{1.06}
\caption{Static merging on LIBERO ($\pi_{0.5}$), underlying Figure~\ref{fig:ablations}b. Success rates (\%) are mean $\pm$ sample SD across three evaluation seeds.}
\label{tab:ablation-static-libero}
\begin{tabularx}{\linewidth}{@{}>{\raggedright\arraybackslash}Xc*{5}{c}@{}}
\toprule
\textbf{Merger} & \textbf{Source} & \textbf{Spatial} & \textbf{Object} & \textbf{Goal} & \textbf{Long} & \textbf{Avg.} \\
\midrule
TIES & S & $66.7\pm3.1$ & $65.3\pm2.1$ & $46.3\pm0.6$ & $29.3\pm4.7$ & $51.9\pm2.1$ \\
WUDI & S & $56.3\pm4.6$ & $62.3\pm2.1$ & $35.7\pm1.2$ & $12.7\pm0.6$ & $41.8\pm0.8$ \\
RegMean++ & S & $73.7\pm2.3$ & $75.3\pm4.6$ & $45.0\pm3.6$ & $15.0\pm3.6$ & $52.3\pm1.0$ \\
\midrule
TIES & G & $77.3\pm1.2$ & $74.7\pm2.3$ & $66.3\pm2.5$ & $50.7\pm6.8$ & $67.3\pm1.6$ \\
WUDI & G & $65.3\pm1.5$ & $61.3\pm2.3$ & $65.0\pm2.6$ & $43.7\pm3.8$ & $58.8\pm1.8$ \\
RegMean++ & G & $86.3\pm5.5$ & $84.7\pm3.5$ & $66.0\pm5.2$ & $55.7\pm3.5$ & $73.2\pm3.4$ \\
\bottomrule
\end{tabularx}
\end{table}

G-isometry improves average static-merging success for each of the three
operators on both benchmarks, showing consistent gains across the evaluated
merging configurations.

\FloatBarrier
\Needspace{8\baselineskip}
\subsubsection{Comparison with Parameter-Space Orthogonal Regularization}
\label{app:regularization-comparison}

We compare G-isometry with no regularization and OrthoReg, a parameter-space
orthogonal regularizer~\citep{orthoreg}. OrthoReg uses its official configuration,
with the expert training budget and RegMean++ merging setup matched across
regularization conditions. G-isometry uses the settings in
Appendix~\ref{app:training-method-settings}. Table~\ref{tab:regularization-comparison}
reports the unmerged task-specific experts, static merges, and merges refined
with OPD for each condition.

\begin{table}[H]
\centering
\fontsize{9}{10.4}\selectfont\setlength{\tabcolsep}{3pt}\renewcommand{\arraystretch}{1.06}
\caption{Regularization comparison across the three benchmarks. Entries are success rates (\%). Each regularization condition is evaluated before merging, after RegMean++ merging, and after subsequent OPD.}
\label{tab:regularization-comparison}
\begin{tabularx}{\linewidth}{@{}l>{\raggedright\arraybackslash}X*{3}{c}@{}}
\toprule
\textbf{Regularization} & \textbf{Evaluated model} & \textbf{SIMPLE} & \textbf{LIBERO} & \textbf{LIBERO-PRO} \\
\midrule
None & Unmerged task-specific experts & 96.7 & 99.5 & 53.1 \\
None & RegMean++ merge & 3.3 & 52.3 & 21.4 \\
None & RegMean++ merge + OPD & 23.3 & 95.4 & 63.3 \\
\midrule
OrthoReg & Unmerged task-specific experts & 100.0 & 99.0 & 52.6 \\
OrthoReg & RegMean++ merge & 23.3 & 68.2 & 32.4 \\
OrthoReg & RegMean++ merge + OPD & 43.3 & 96.6 & 65.7 \\
\midrule
G-isometry & Unmerged task-specific experts & 100.0 & 99.3 & 53.7 \\
G-isometry & RegMean++ merge & 40.0 & 73.2 & 36.9 \\
G-isometry & RegMean++ merge + OPD & 86.7 & 97.1 & 67.3 \\
\bottomrule
\end{tabularx}
\end{table}

G-isometry yields higher static-merge and post-OPD performance than the other
tested regularization conditions on all three benchmarks.
Table~\ref{tab:regularization-simple} further shows more balanced skill retention
on SIMPLE: after OPD, all three tasks achieve at least eight successes in ten
trials. These results support the benefit of the tested response-space
regularizer for expert merging and subsequent refinement.

\begin{table}[H]
\centering
\fontsize{9}{10.4}\selectfont\setlength{\tabcolsep}{3pt}\renewcommand{\arraystretch}{1.06}
\caption{Per-task SIMPLE results for the regularization comparison. Task entries are successful episodes out of ten. Avg. is the task-macro success rate (\%).}
\label{tab:regularization-simple}
\begin{tabularx}{\linewidth}{@{}l>{\raggedright\arraybackslash}X*{4}{c}@{}}
\toprule
\textbf{Regularization} & \textbf{Evaluated model} & \textbf{BendPick} & \textbf{CloseDoor} & \textbf{OpenOven} & \textbf{Avg.} \\
\midrule
None & RegMean++ merge & 0/10 & 0/10 & 1/10 & 3.3 \\
None & RegMean++ merge + OPD & 0/10 & 6/10 & 1/10 & 23.3 \\
\midrule
OrthoReg & RegMean++ merge & 0/10 & 3/10 & 4/10 & 23.3 \\
OrthoReg & RegMean++ merge + OPD & 1/10 & 7/10 & 5/10 & 43.3 \\
\midrule
G-isometry & RegMean++ merge & 3/10 & 5/10 & 4/10 & 40.0 \\
G-isometry & RegMean++ merge + OPD & 8/10 & 10/10 & 8/10 & 86.7 \\
\bottomrule
\end{tabularx}
\end{table}

\FloatBarrier
\Needspace{8\baselineskip}
\subsubsection{Sensitivity to Regularization Strength}
\label{app:regularization-strength}

We vary the SIMPLE regularization strength over $\beta\in\{0,1,2,4\}$,
with default $\beta_0=2$, while retaining the other expert-training settings and
the RegMean++ merging configuration. For each nonzero strength, OPD starts from
that condition's merged model and uses experts trained at the same strength as
teachers. The $\beta=0$ OPD entry is the standard-merge reference from
Table~\ref{tab:ablation-opd-initialization}, whose initialization comparison
keeps teachers fixed. Each evaluated model uses ten trials per task.

\begin{table}[H]
\centering
\fontsize{9}{10.4}\selectfont\setlength{\tabcolsep}{3pt}\renewcommand{\arraystretch}{1.06}
\caption{Regularization-strength comparison on SIMPLE. All entries are success rates (\%). Expert columns evaluate the unmerged task-specific models. Static-merge and post-OPD columns average equally over the three tasks. The default is $\beta_0=2$.}
\label{tab:regularization-strength}
\begin{tabularx}{\linewidth}{@{}>{\raggedright\arraybackslash}X*{6}{c}@{}}
\toprule
\textbf{$\beta/\beta_0$} & \shortstack{\textbf{Expert}\\\textbf{BendPick}} & \shortstack{\textbf{Expert}\\\textbf{CloseDoor}} & \shortstack{\textbf{Expert}\\\textbf{OpenOven}} & \shortstack{\textbf{Expert}\\\textbf{avg.}} & \shortstack{\textbf{Static merge}\\\textbf{avg.}} & \shortstack{\textbf{Post-OPD}\\\textbf{avg.}} \\
\midrule
0 (Standard FT) & 90.0 & 100.0 & 100.0 & 96.7 & 3.3 & 23.3 \\
0.5 & 100.0 & 100.0 & 100.0 & 100.0 & 20.0 & 43.3 \\
1.0 (default) & 100.0 & 100.0 & 100.0 & 100.0 & 40.0 & 86.7 \\
2.0 & 90.0 & 100.0 & 90.0 & 93.3 & 46.7 & 86.7 \\
\bottomrule
\end{tabularx}
\end{table}

\begin{table}[H]
\centering
\fontsize{9}{10.4}\selectfont\setlength{\tabcolsep}{3pt}\renewcommand{\arraystretch}{1.06}
\caption{Per-task SIMPLE results after OPD at each regularization strength. Task entries are successful episodes out of ten. Avg. is the task-macro success rate (\%).}
\label{tab:regularization-strength-simple}
\begin{tabularx}{\linewidth}{@{}>{\raggedright\arraybackslash}X*{4}{c}@{}}
\toprule
\textbf{$\beta/\beta_0$} & \textbf{BendPick} & \textbf{CloseDoor} & \textbf{OpenOven} & \textbf{Avg.} \\
\midrule
0 (Standard FT) & 0/10 & 6/10 & 1/10 & 23.3 \\
0.5 & 3/10 & 5/10 & 5/10 & 43.3 \\
1.0 (default) & 8/10 & 10/10 & 8/10 & 86.7 \\
2.0 & 7/10 & 10/10 & 9/10 & 86.7 \\
\bottomrule
\end{tabularx}
\end{table}

Half the default strength improves static merging over standard fine-tuning
but achieves lower post-OPD success than the default. Doubling the strength
raises static-merge success from 40.0\% to 46.7\%, while the observed expert
average falls from 100.0\% to 93.3\%. Both strengths reach 86.7\% after OPD.
The task counts in Table~\ref{tab:regularization-strength-simple} show different
skill-level outcomes despite the equal final averages. Within this tested range,
the default retains the observed standalone expert performance while achieving
strong multi-skill integration.

\FloatBarrier
\Needspace{8\baselineskip}
\subsection{Distillation Design}
\label{app:distillation-design}

We next examine the roles of student initialization, environment observations,
and teacher supervision in post-merging refinement.

\subsubsection{Student Initialization}
Table~\ref{tab:ablation-opd-initialization} reports final SIMPLE outcomes. B starts from the pretrained base, while S and G start from the corresponding merged models. This initialization ablation uses one training run per condition. Repeated post-training runs for Joint FT(G) and GeoMerge-OPD(G) are reported separately in Appendix~\ref{app:training-seeds}.

\begin{table}[H]
\centering
\fontsize{9}{10.4}\selectfont\setlength{\tabcolsep}{3pt}\renewcommand{\arraystretch}{1.06}
\caption{OPD initialization on SIMPLE ($\Psi_0$), underlying Figure~\ref{fig:ablations}c. The teachers are unchanged across initializations. Task entries are successful episodes out of ten. Avg. is the task-macro success rate (\%).}
\label{tab:ablation-opd-initialization}
\begin{tabularx}{\linewidth}{@{}>{\raggedright\arraybackslash}Xc*{4}{c}@{}}
\toprule
\textbf{Initialization} & \textbf{Code} & \textbf{BendPick} & \textbf{CloseDoor} & \textbf{OpenOven} & \textbf{Avg.} \\
\midrule
Pretrained base (direct OPD) & B & 1/10 & 8/10 & 0/10 & 30.0 \\
Standard merged model & S & 0/10 & 6/10 & 1/10 & 23.3 \\
G-isometry merged model & G & 8/10 & 10/10 & 8/10 & 86.7 \\
\bottomrule
\end{tabularx}
\end{table}

With the teachers fixed, the G-regularized initialization yields 86.7\%
success, compared with 30.0\% from the pretrained base and 23.3\% from the
standard merge under the evaluated OPD setup.

\FloatBarrier
\Needspace{8\baselineskip}
\subsubsection{Environment Observation Source}
\label{app:observation-source}

We replace student-collected environment observations with fixed demonstration
observations while retaining the same G-regularized initialization, task-specific
teachers, and velocity-field loss. Both settings use student-generated
action-flow trajectories, the same gradient computation, and matched
teacher-supervision and optimization budgets. Thus, this comparison changes the
environment observation source while retaining student-generated action-flow
states in both conditions.

\begin{table}[H]
\centering
\fontsize{9}{10.4}\selectfont\setlength{\tabcolsep}{3pt}\renewcommand{\arraystretch}{1.06}
\caption{Observation-source ablation. Success rates are reported in percent. Improvements are in percentage points (pp). Both conditions supervise intermediate velocity fields along student-generated action-flow trajectories.}
\label{tab:observation-source}
\begin{tabularx}{\linewidth}{@{}>{\raggedright\arraybackslash}X>{\raggedright\arraybackslash}X*{3}{c}@{}}
\toprule
\textbf{Environment observations} & \textbf{Teacher supervision} & \textbf{SIMPLE} & \textbf{LIBERO} & \textbf{LIBERO-PRO} \\
\midrule
Fixed demonstrations & Intermediate velocity fields & 66.7 & 94.7 & 62.8 \\
Student-collected & Intermediate velocity fields & 86.7 & 97.1 & 67.3 \\
\midrule
Improvement (pp) & --- & $+20.0$ & $+2.4$ & $+4.5$ \\
\bottomrule
\end{tabularx}
\end{table}

Under these controls, student-collected observations improve success by
20.0, 2.4, and 4.5 percentage points on SIMPLE, LIBERO, and LIBERO-PRO,
respectively. The results support the contribution of distilling on environment
states visited by the student to the shared policy's closed-loop performance.

\FloatBarrier
\Needspace{8\baselineskip}
\subsubsection{Final-Action and Velocity-Field Supervision}
\label{app:teacher-supervision}

We compare velocity-field distillation with a final-action baseline that trains
the student by flow matching using teacher-generated actions as labels. Both
settings use the same G-regularized initialization and task-specific teachers,
with matched environment-interaction and optimizer-update budgets.
Table~\ref{tab:teacher-supervision} compares learning from the teachers' final
actions with teacher velocity-field supervision at the student's intermediate
action-flow states.

\begin{table}[H]
\centering
\fontsize{9}{10.4}\selectfont\setlength{\tabcolsep}{3pt}\renewcommand{\arraystretch}{1.06}
\caption{Final-action versus intermediate velocity-field supervision. Success rates are in percent. Improvements are in percentage points (pp).}
\label{tab:teacher-supervision}
\begin{tabularx}{\linewidth}{@{}>{\raggedright\arraybackslash}X>{\raggedright\arraybackslash}X*{3}{c}@{}}
\toprule
\textbf{Teacher target} & \textbf{Student objective} & \textbf{SIMPLE} & \textbf{LIBERO} & \textbf{LIBERO-PRO} \\
\midrule
Final actions & Flow matching with teacher-generated action labels & 73.3 & 95.7 & 65.8 \\
Intermediate velocity fields & Velocity-field distillation & 86.7 & 97.1 & 67.3 \\
\midrule
Improvement (pp) & --- & $+13.3$ & $+1.4$ & $+1.5$ \\
\bottomrule
\end{tabularx}
\end{table}

Velocity-field supervision achieves higher success on all three benchmarks
under these matched conditions, supporting the benefit of supervising the
student's intermediate action-generation states for closed-loop performance.

\FloatBarrier
\Needspace{8\baselineskip}
\subsection{Variability Across Post-Training Seeds}
\label{app:training-seeds}

We repeat Joint FT and GeoMerge-OPD with two additional independent training
seeds. Seed~1 is the original run, and Seeds~2 and~3 are the additional runs.
For G-initialized methods, the experts and merged initialization are fixed
across runs, and only post-training is repeated. B denotes the pretrained base,
and G denotes the merge of G-isometry-regularized experts.

\begin{table}[H]
\centering
\fontsize{9}{10.4}\selectfont\setlength{\tabcolsep}{3pt}\renewcommand{\arraystretch}{1.06}
\caption{Per-training-seed success rates (\%) for Joint FT and GeoMerge-OPD. Mean and sample SD are computed across the three observed runs. For SIMPLE, statistics are computed from the underlying success counts before rounding.}
\label{tab:training-seeds}
\begin{tabularx}{\linewidth}{@{}l>{\raggedright\arraybackslash}X*{4}{c}@{}}
\toprule
\textbf{Benchmark} & \textbf{Method / initialization} & \textbf{Seed 1} & \textbf{Seed 2} & \textbf{Seed 3} & \textbf{Mean $\pm$ SD} \\
\midrule
SIMPLE & Joint FT(G) & 73.3 & 66.7 & 63.3 & $67.8\pm5.1$ \\
SIMPLE & GeoMerge-OPD(G) & 86.7 & 86.7 & 83.3 & $85.6\pm1.9$ \\
\midrule
LIBERO & Joint FT(B) & 96.2 & 94.2 & 96.5 & $95.6\pm1.3$ \\
LIBERO & Joint FT(G) & 95.2 & 95.4 & 96.9 & $95.8\pm0.9$ \\
LIBERO & GeoMerge-OPD(G) & 97.1 & 95.9 & 97.7 & $96.9\pm0.9$ \\
\midrule
LIBERO-PRO & Joint FT(B) & 64.5 & 63.7 & 63.6 & $63.9\pm0.5$ \\
LIBERO-PRO & Joint FT(G) & 64.9 & 64.8 & 65.3 & $65.0\pm0.3$ \\
LIBERO-PRO & GeoMerge-OPD(G) & 67.3 & 66.5 & 68.3 & $67.4\pm0.9$ \\
\bottomrule
\end{tabularx}
\end{table}

GeoMerge-OPD exceeds the evaluated Joint FT baselines on every benchmark at each
of the three training seeds. Relative to Joint FT(G) from the same merged
initialization, the three-run mean improves by 17.8, 1.1, and 2.4 percentage
points on SIMPLE, LIBERO, and LIBERO-PRO, respectively. On SIMPLE, the per-task counts in
Table~\ref{tab:training-seeds-simple} show that this advantage is accompanied
by consistently stronger BendPick retention and balanced performance across
the three skills.

\begin{table}[H]
\centering
\fontsize{9}{10.4}\selectfont\setlength{\tabcolsep}{3pt}\renewcommand{\arraystretch}{1.06}
\caption{Per-task SIMPLE outcomes across training seeds. Task entries are successful episodes out of ten. Avg. is the task-macro success rate (\%). $\Delta$ is the improvement over Joint FT(G) for the corresponding training seed, in percentage points.}
\label{tab:training-seeds-simple}
\begin{tabularx}{\linewidth}{@{}c>{\raggedright\arraybackslash}X*{5}{c}@{}}
\toprule
\textbf{Seed} & \textbf{Method} & \textbf{BendPick} & \textbf{CloseDoor} & \textbf{OpenOven} & \textbf{Avg.} & \textbf{$\Delta$ (pp)} \\
\midrule
1 & Joint FT(G) & 5/10 & 9/10 & 8/10 & 73.3 & --- \\
1 & GeoMerge-OPD(G) & 8/10 & 10/10 & 8/10 & 86.7 & $+13.3$ \\
\midrule
2 & Joint FT(G) & 4/10 & 8/10 & 8/10 & 66.7 & --- \\
2 & GeoMerge-OPD(G) & 8/10 & 9/10 & 9/10 & 86.7 & $+20.0$ \\
\midrule
3 & Joint FT(G) & 4/10 & 8/10 & 7/10 & 63.3 & --- \\
3 & GeoMerge-OPD(G) & 7/10 & 10/10 & 8/10 & 83.3 & $+20.0$ \\
\bottomrule
\end{tabularx}
\end{table}

\paragraph{Experimental scope and evaluation limitations.}
The repeated-seed results for G-initialized methods are conditional on fixed
experts and a fixed merged initialization. SIMPLE uses ten evaluation trials
per task per training seed. The reported SDs describe variation across the
observed runs, subject to this finite evaluation budget. They are not confidence
intervals for evaluation success probabilities. The regularization comparisons, observation-source and teacher-supervision
ablations, and strength sweep do not include corresponding repeated
training-seed estimates. These experiments cover $\Psi_0$ on SIMPLE and
$\pi_{0.5}$ on LIBERO and LIBERO-PRO under the evaluated configurations.
Generalization to other backbones, tasks, and training budgets remains untested.

\FloatBarrier
\Needspace{10\baselineskip}
\section{Response Geometry and Behavioral Diagnostics}
\label{app:mechanism-diagnostics}

\noindent\textbf{Comparison.}
We diagnose response geometry and merged-policy behavior for two-expert
RegMean++ merging of the $\Psi_0$ BendPick (B) and CloseDoor (C) experts.
Table~\ref{tab:mechanism-diagnostics} compares standard fine-tuning, OrthoReg,
and G-isometry. The original standard/G-isometry comparison uses the same
RegMean++ configuration in both conditions. This two-task diagnostic complements
the three-task SIMPLE integration experiments.

\noindent\textbf{Fixed inputs.}
For the original standard/G-isometry comparison, we select 20 held-out reference trajectories per task and uniformly sample 16 valid replanning observations per trajectory. At the midpoint of each of eight equal-width noise bins, we use three fixed noise samples. Conditions share images, instructions, proprioception, normalization, and real-time chunking (RTC) prefixes and delays. We collect layer inputs $x$ from the common pretrained base at the constrained linear layers and estimate the full input second moment $G=\mathbb E[xx^\top]$ for each task, layer, and bin. Invalid tokens, padded frames, and committed action prefixes are masked. The layer set is fixed in advance.

\noindent\textbf{Response geometry.}
At each layer and bin, define the full response matrix $M_q^{\mathrm{full}}=\Delta W_qG_q^{1/2}$. The training regularizer uses retained activation directions. Each model uses its own spectrum for $C_5$. For cosine evaluation in the original two conditions, fix the five leading right singular vectors $v_j$ of the standard BendPick response matrix. Under the common BendPick reference, let $r_{q,j}=\Delta W_qG_B^{1/2}v_j$ and compute
\begin{equation}
C_5(q)=\frac{\sum_{j=1}^{5}\sigma_j^2(M_q^{\mathrm{full}})}{\|M_q^{\mathrm{full}}\|_F^2},
\qquad
c_j=\frac{\langle r_{B,j},r_{C,j}\rangle}{\|r_{B,j}\|_2\|r_{C,j}\|_2}.
\label{app:eq:diagnostic-metrics}
\end{equation}
Cosines are first reduced by their median over the five directions. Both this statistic and $C_5$ are averaged equally across bins, then reduced by their median across layers. The aggregated cosine is denoted $c_B$. Near-zero responses are marked undefined using a prespecified threshold. We retain common valid directions and record their count. The denominator of $C_5$ is the full response energy.

\noindent\textbf{Behavioral evaluation.}
The original standard/G-isometry diagnostic uses 20 paired scenarios per task, restoring the complete initial environment state and fixing policy noise, controller, sampler, RTC, and success criteria. BendPick translation-command magnitude $V_{xy}$ is the episode time average of $\sqrt{v_x^2+v_y^2}$ after denormalization, averaged equally across episodes. It includes executed commands, excludes warm-up and unexecuted trailing chunks, and omits turning flags and target heading. Success counts on both tasks help identify overall action shrinkage or early failure. Table~\ref{tab:mechanism-diagnostics} combines the previously recorded standard/G-isometry results with the additional OrthoReg comparison. Its success counts use ten trials per policy per task.

\begin{table}[ht!]
\centering
\fontsize{9}{10.5}\selectfont
\setlength{\tabcolsep}{4pt}
\renewcommand{\arraystretch}{1.07}
\caption{Response geometry and two-expert RegMean++ merging results, including the additional OrthoReg comparison. Success counts in (b) use ten trials per policy per task.}
\label{tab:mechanism-diagnostics}
\begin{tabular*}{\linewidth}{@{\extracolsep{\fill}}lccc@{}}
\toprule
\multicolumn{4}{@{}l}{\textbf{(a) Expert-update responses}}\\
\midrule
\textbf{Expert fine-tuning} & \textbf{BendPick $C_5$ (\%)} & \textbf{CloseDoor $C_5$ (\%)} & \textbf{Cosine $c_B$}\\
\midrule
Standard FT & 78.1 & 93.3 & $-0.73$\\
OrthoReg FT & 69.7 & 75.1 & $-0.48$\\
G-isometry FT & 17.6 & 14.3 & $-0.31$\\
\midrule
\multicolumn{4}{@{}l}{\textbf{(b) Single-task reference and two-expert merging}}\\
\midrule
\textbf{Policy} & \textbf{BendPick $V_{xy}$} & \textbf{BendPick} & \textbf{CloseDoor}\\
\midrule
Standard FT BendPick expert & 0.0013 & 9/10 & 0/10\\
Standard FT + RegMean++ & 0.45 & 0/10 & 5/10\\
OrthoReg FT + RegMean++ & 0.19 & 1/10 & 5/10\\
G-isometry FT + RegMean++ & 0.034 & 3/10 & 7/10\\
\bottomrule
\end{tabular*}
\end{table}

Relative to standard FT and OrthoReg, G-isometry yields lower response-energy concentration, a less negative response cosine, and smaller BendPick translation commands, together with higher success counts in this two-task setting. This diagnostic examines correspondence between response structure and behavioral changes. It does not by itself establish that spectral concentration causes task failure.

\FloatBarrier
\Needspace{10\baselineskip}
\section{Theory}
\label{app:theory}

This appendix develops four results for GeoMerge-OPD. They concern
interference in high-energy response directions during parameter merging,
the responses controlled by G-isometry, transient task degradation during
joint training, and the link between two-level OPD error and task performance.
The response analysis uses a linear layer with fixed activations.
The behavioral analysis uses the finite-step action sampler in the main text.
Related background includes functional interpretations of task arithmetic
\citep{tangent_task_arithmetic} and multitask gradient interference
\citep{pcgrad}.

Let $R$ be the number of task groups, indexed by $p,q\in\{1,\ldots,R\}$.
A group is one SIMPLE task or one LIBERO suite.
The matrix norms $\|\cdot\|_F$ and $\|\cdot\|_{\mathrm{op}}$ denote the
Frobenius norm and the largest singular value, respectively. Vector norms
without subscripts are Euclidean.
$\langle A,B\rangle_F=\operatorname{tr}(A^\top B)$ denotes the inner product
of matrices with the same shape, $\operatorname{tr}$ denotes the trace, and
$I$ is the identity matrix of the dimension determined by context.

\subsection{Task-Conditioned Responses and Merging Deviations}
\label{app:merge-response}

Fix a layer and a flow time bin. Let $h\in\mathbb R^{d_{\mathrm{in}}}$ be the
layer input, with input and output dimensions $d_{\mathrm{in}}$ and
$d_{\mathrm{out}}$, respectively. The update of expert weights $W_q$ relative
to the shared base $W_0$ is $D_q=W_q-W_0$. The reference input second-moment
matrix for task $q$ is $G_q=\mathbb E_q[hh^\top]$, where $\mathbb E_q$ denotes
expectation over that task's reference input distribution. Retaining $r_q$
directions gives a factor $S_q\in\mathbb R^{d_{\mathrm{in}}\times r_q}$
corresponding to $\widehat G_q=S_qS_q^\top$. All comparisons across experts
use the same task-specific reference factor:
\begin{equation}
 B_{pq}=D_pS_q,\qquad M_q=B_{qq},\qquad
 \mathcal E_q=\|M_q\|_F^2.
 \label{app:eq:response}
\end{equation}
Here, $B_{pq}$ is the response of expert $p$'s update on the reference inputs
of task $q$. The squared singular values of $M_q$ give the response energies
in the retained space. This relation is exact for fixed layer inputs. It does
not account for nonlinearities between layers or parameter-dependent inputs
in the full network.

First consider additive merging $D_{\mathrm m}=\sum_p a_pD_p$, where
$a_p\ge0$ is the merging coefficient of expert $p$. On task $q$, the retained
response error relative to its expert is
\begin{equation}
 F_q=\|(D_{\mathrm m}-D_q)S_q\|_F^2.
 \label{app:eq:merge-error}
\end{equation}
Let $P_q$ be an orthogonal projection in the output space, for example onto
$M_q$'s leading $k$ left singular directions, where $k$ is the number of
selected directions. Define
\begin{equation}
 U_q=P_qM_q,\quad V_q=\sum_{p\ne q}a_pP_qB_{pq},\quad
 \rho_q=\frac{\|U_q\|_F^2}{\mathcal E_q},\quad
 b_q=\frac{\|V_q\|_F}{\|U_q\|_F},\quad
 c_q=\frac{\langle U_q,V_q\rangle_F}{\|U_q\|_F\|V_q\|_F}.
 \label{app:eq:dominant-geometry}
\end{equation}
The matrix $U_q$ is the task's own response in the selected directions, and
$V_q$ is the merged response of other experts in those same directions.
The quantity $\rho_q$ is the energy fraction carried by these directions,
$b_q$ is the ratio of the other experts' response magnitude to the task's own
response magnitude, and $c_q$ is their directional cosine. Assume
$\mathcal E_q>0$ and $U_q\ne0$. If $V_q=0$, set $b_q=0,c_q=0$. We use a
matrix Frobenius cosine. Medians of layerwise or directionwise cosines differ
from this aggregate quantity.

\begin{theoryprop}[Merging deviation under opposing responses]
\label{app:prop:merge}
Under opposing interference, concentrating more response energy in the
affected dominant directions increases the merging deviation in those
directions. Specifically, fix $\mathcal E_q,a_q,b_q,c_q$. If
$0\le a_q\le1$, $b_q>0$, and $c_q\le0$, then
\begin{equation}
\begin{split}
 \|P_q(D_{\mathrm m}-D_q)S_q\|_F^2
 &=\rho_q\mathcal E_q
 \big[(1-a_q)^2+b_q^2-2(1-a_q)b_qc_q\big]\\
 &\le F_q,
\end{split}
\label{app:eq:merge-dominant}
\end{equation}
where the right-hand side of the equality increases strictly with the energy
fraction $\rho_q$.
\end{theoryprop}

Equation~\eqref{app:eq:merge-dominant} separates the deviation in the dominant
directions into three parts: the change in the task's own response caused by
its merging weight, the response introduced by other experts, and their
interaction. When $c_q<0$ and $a_q<1$, the last term is positive, so opposing
responses further increase the deviation. The factor $\rho_q\mathcal E_q$
determines how much task response these affected directions carry.

\begin{proof}[Proof]
The projected error is $(a_q-1)U_q+V_q$. Expanding its squared norm gives
\[
 (1-a_q)^2\|U_q\|_F^2+\|V_q\|_F^2
 -2(1-a_q)\langle U_q,V_q\rangle_F.
\]
Substituting Eq.~\eqref{app:eq:dominant-geometry} yields the equality. Since
$P_q$ is an orthogonal projection, the full error additionally contains the
nonnegative squared norm in the orthogonal complement, yielding the
inequality. The final claim follows from the signs of the bracketed terms.
\end{proof}

This result describes the combined effect of concentration and opposing
responses. Energy may concentrate because a skill requires only a few control
directions. Concentration alone is not harmful. For BendPick, if responses
from other experts cancel necessary adjustments in its dominant response
space or shift the output toward the control requirements of other tasks,
the resulting deviation lies in directions on which this task relies
heavily. Internal response directions do not directly correspond to
translation command directions. Subsequent network layers and environment
dynamics also mediate the relationship between layer response deviations and
closed-loop task failure.

Equation~\eqref{app:eq:merge-dominant} applies to additive merging. For merging
rules with parameter masks, matrix weighting, or other nonadditive operations,
Eq.~\eqref{app:eq:merge-error} should be evaluated using the actual
$D_{\mathrm m}$. If $D_{\mathrm m}=\sum_pa_pD_p+E_{\mathrm m}$ and $Q_q$
denotes the equality value in Eq.~\eqref{app:eq:merge-dominant}, the triangle
inequality still gives
\[
 F_q\ge\big[\max\{0,\sqrt{Q_q}-\|P_qE_{\mathrm m}S_q\|_F\}\big]^2.
\]
Additional merging operations can therefore also correct the original
additive deviation.

\subsection{Gain and Coverage Controlled by G-isometry}
\label{app:g-isometry}

The regularizer in the main text acts on each expert's own $M_p=D_pS_p$:
\begin{equation}
 R_p=\frac1{r_p}\|M_p^\top M_p-I_{r_p}\|_F^2,
 \qquad \delta_p=\sqrt{r_pR_p}.
 \label{app:eq:regularizer}
\end{equation}
It bounds response gain weighted by task activations. We next relate this
property to responses on other tasks without assuming identical activation
subspaces across tasks.

To express activation coverage across tasks, assume each $S_p$ has full
column rank and define
\begin{equation}
 A_{pq}=S_p^\dagger S_q,\qquad
 T_{pq}=(I-S_pS_p^\dagger)S_q,\qquad
 u_{pq}=\sqrt{1+\delta_p}\|A_{pq}\|_F+\|D_pT_{pq}\|_F.
 \label{app:eq:coverage}
\end{equation}
Here, $\dagger$ denotes the Moore--Penrose pseudoinverse. The matrix $A_{pq}$
represents the part of task $q$ covered by expert $p$'s retained directions,
$T_{pq}$ represents the uncovered part, and $u_{pq}$ bounds the response
magnitude contributed by both parts.

\begin{theoryprop}[Response gain under cross-task coverage]
\label{app:prop:regularizer}
G-isometry limits response amplification in covered directions and thereby
bounds the merging deviation. Under the preceding setup,
\begin{equation}
 \|D_pS_q\|_F\le u_{pq},\qquad
 \sqrt{F_q}\le |a_q-1|\|M_q\|_F+
                  \sum_{p\ne q}a_pu_{pq}.
 \label{app:eq:merge-envelope}
\end{equation}
\end{theoryprop}

The first inequality in Eq.~\eqref{app:eq:merge-envelope} bounds expert $p$'s
response magnitude on task $q$. The second sums these magnitudes to bound
the merged model's deviation from the task's own expert. The regularizer
controls the gain of the covered part through $\sqrt{1+\delta_p}$, while the
uncovered response $\|D_pT_{pq}\|_F$ remains an explicit term.

\begin{proof}[Proof]
The operator norm of a symmetric matrix is bounded by its Frobenius norm, so
$\|M_p^\top M_p-I\|_{\mathrm{op}}\le\delta_p$.
Let the eigenvalues of $M_p^\top M_p$ be $\lambda_j$. Then
$|\lambda_j-1|\le\delta_p$, giving
\begin{equation}
 \|M_p\|_{\mathrm{op}}\le\sqrt{1+\delta_p}.
 \label{app:eq:gain-upper}
\end{equation}
If $\delta_p<1$, each retained singular value $\sigma_j(M_p)$ also satisfies
$\sqrt{1-\delta_p}\le\sigma_j(M_p)\le\sqrt{1+\delta_p}$.
Also, $S_q=S_pA_{pq}+T_{pq}$, hence
\[
 D_pS_q=M_pA_{pq}+D_pT_{pq}.
\]
The inequality $\|MA\|_F\le\|M\|_{\mathrm{op}}\|A\|_F$ and the triangle
inequality give the first response bound. Applying the triangle inequality
again to
$(D_{\mathrm m}-D_q)S_q=(a_q-1)M_q+\sum_{p\ne q}a_pD_pS_q$
gives the second bound.
\end{proof}

Robot tasks share vision, proprioception, and action generation modules,
making shared activation directions plausible. Different instructions,
objects, and manipulation stages can also activate directions outside the
retained space. Accordingly, Eq.~\eqref{app:eq:merge-envelope} retains
$\|D_pT_{pq}\|_F$ rather than assuming unconditional coverage across tasks.
Reducing $R_p$ tightens the gain factor for the covered component. Across
training runs, $S_p$, relative directions, magnitudes, and uncovered
components may also change.

The spectral meaning of the regularizer can also be expressed by an exact
decomposition. Let $\lambda_j$ be the eigenvalues of $M_p^\top M_p$ and let
$\bar\lambda=r_p^{-1}\sum_j\lambda_j$. Then
\[
 R_p=(\bar\lambda-1)^2+\frac1{r_p}\sum_j(\lambda_j-\bar\lambda)^2.
\]
When $\delta_p<1$, the energy fraction in the leading $k$ response directions
is at most $\min\{1,k(1+\delta_p)/(r_p(1-\delta_p))\}$, because the numerator
is at most $k(1+\delta_p)$ and the denominator is at least $r_p(1-\delta_p)$.
This worst-case fraction bound requires $\delta_p<1$. The upper bound in
Eq.~\eqref{app:eq:gain-upper} does not.

Controlled gain and compatibility across tasks are distinct conditions. For
example, with $S_1=S_2=I$, $D_1=I$, and $D_2=-I$, both regularizers are zero,
yet equal-weight merging produces a zero update that still deviates from
each expert. G-isometry directly controls directional imbalance in update
responses. See \citet{orthoreg} for related orthogonal regularization in
parameter space. It does not constrain the entire spectrum of the full
mapping $W_0+D_p$, whose Gram matrix also contains base and cross terms. If
the full input second-moment matrix is $G_q=\widehat G_q+E_q$, with
$E_q\succeq0$, the full layer error additionally contains
$\operatorname{tr}((D_{\mathrm m}-D_q)E_q(D_{\mathrm m}-D_q)^\top)$.

\subsection{When Directional Imbalance Affects Joint Training}
\label{app:joint-training}

Joint training determines its next update at a common parameter point.
Let $\theta$ denote the shared parameters, $L_q(\theta)$ the task loss, and
$g_q=\nabla_\theta L_q$ its gradient. Task weights satisfy $\alpha_q>0$ and
$\sum_q\alpha_q=1$. The average loss and gradient are
$\bar L=\sum_q\alpha_qL_q$ and $\bar g=\sum_q\alpha_qg_q$, respectively.
For fixed sampling distributions, if $L_q$ is twice continuously
differentiable, using learning rate $\eta>0$ for the update
$\theta^+=\theta-\eta\bar g$ gives
\begin{equation}
 L_q(\theta-\eta\bar g)-L_q(\theta)
 =-\eta\left[\alpha_q\|g_q\|^2+
       \sum_{p\ne q}\alpha_p g_q^\top g_p\right]+O(\eta^2).
 \label{app:eq:joint-first-order}
\end{equation}
Inside the brackets, $\alpha_q\|g_q\|^2$ is the task's own improvement term,
and the remaining inner products describe the effects of other tasks.
First-order degradation occurs only when the weighted negative terms
outweigh this improvement. The term $O(\eta^2)$ contains higher-order effects
that scale quadratically with the learning rate. This distinction is
consistent with analyses of multitask gradient interference \citep{pcgrad}.

Assume that all inputs and targets below have finite second moments.
Consider a fixed-feature linear predictor with current weights $W$. Let
$h_q,y_q$ be the input and supervision target for task $q$, respectively,
and let $e_q=Wh_q-y_q$ be the prediction residual. The squared-loss gradients
satisfy
\begin{equation}
 g_q=\mathbb E_q[e_qh_q^\top],\qquad
 \langle g_q,g_p\rangle_F
 =\mathbb E_{q,p}[(e_q^\top e_p)(h_q^\top h_p)].
 \label{app:eq:gradient-overlap}
\end{equation}
The two tasks are sampled independently on the right-hand side. This
identity follows by expanding the Frobenius inner product of the two
expectations. For a shared linear layer in a deep network, $e_q$ can be
replaced by the backpropagated derivative at that layer's output. Hidden
response directions therefore affect task losses through the subsequent
network.

In robot tasks, the following conditions can jointly produce the negative
effect in Eq.~\eqref{app:eq:joint-first-order}. First, two tasks use shared
visual, proprioceptive, or locomotion features, and task conditioning has not
yet fully separated the relevant gradients. Second, the current predictions
require different corrections. If the model commands excessive motion during
stationary grasping but insufficient motion when approaching a cabinet door,
the two tasks require opposite local corrections to the shared output.
The comparison concerns corrections relative to current predictions.
Zero and nonzero motion targets are not inherently conflicting. Third, a
larger residual magnitude, sampling weight, or local sensitivity can make
one task's weighted contribution dominant. Finally, directions that
distinguish tasks may remain available but be relatively weak, requiring
more updates to compensate for a shift along the shared direction. The
following realizable model makes these conditions concrete.

Consider two tasks sharing parameters $w\in\mathbb R^2$, with inputs
$h_1=(A,1)^\top$ and $h_2=(A,-1)^\top$ and targets $y_1=1,y_2=-2$.
Here, $A>\sqrt3$ represents the strength of the shared direction relative to
the direction that distinguishes the tasks. Define
$L_q(w)=\tfrac12(w^\top h_q-y_q)^2$ and apply gradient descent to the equally
weighted average loss from $w_0=0$. Let $w_t$ denote the parameters after
$t$ updates. Both tasks admit the common zero-loss solution
$w^\star=(-1/(2A),3/2)^\top$.

\begin{theoryprop}[Transient degradation of jointly realizable tasks]
\label{app:prop:joint}
Even when two tasks can be learned simultaneously, joint training can
initially worsen one task while the average loss keeps decreasing. In the
preceding model, take $\eta=A^{-2}$. For every integer $t\ge1$,
\begin{equation}
 L_1(w_t)=L_2(w_t)=\frac98(1-A^{-2})^{2t}.
 \label{app:eq:joint-transient}
\end{equation}
After the first step, task one's loss exceeds its initial value $1/2$.
It then recovers gradually and converges to zero together with task two.
\end{theoryprop}

The factor $(1-A^{-2})^{2t}$ in Eq.~\eqref{app:eq:joint-transient} describes
the gradual decrease in error during training. A stronger shared direction
means a larger $A$, bringing this decay factor closer to $1$. The direction
that distinguishes the tasks then takes longer to compensate for the
initial shift. Task one remains worse than at initialization over the
interval
\begin{equation}
 1\le t<\frac{\log(3/2)}{-\log(1-A^{-2})}.
 \label{app:eq:duration}
\end{equation}
Let the mixed input second-moment matrix be
$\bar G=\tfrac12(h_1h_1^\top+h_2h_2^\top)=\operatorname{diag}(A^2,1)$.
Its condition number $\kappa(\bar G)$, the ratio of its largest to smallest
eigenvalue, equals $A^2$. As $A\to\infty$, the threshold in
Eq.~\eqref{app:eq:duration} is approximately
$\log(3/2)\kappa(\bar G)$. Thus, greater directional imbalance can prolong
this transient degradation. All occurrences of $\log$ denote natural
logarithms.

\begin{proof}[Proof]
Substituting $w^\star$ directly gives the targets $1$ and $-2$.
At initialization,
\[
 g_1=(-A,-1)^\top,\quad g_2=(2A,-2)^\top,\quad
 \bar g=(A/2,-3/2)^\top,\quad
 g_1^\top\bar g=(3-A^2)/2.
\]
For any $w$, the squared loss gives the exact change
\begin{equation}
 L_q(w-\eta\bar g)-L_q(w)
 =-\eta g_q^\top\bar g+
        \frac{\eta^2}{2}(h_q^\top\bar g)^2.
 \label{app:eq:joint-exact}
\end{equation}
At initialization, when $A^2>3$, the first-order term for task one is strictly
positive and the second-order term is nonnegative. The change in the average loss is
$-\eta\|\bar g\|^2+\tfrac{\eta^2}{2}\bar g^\top\bar G\bar g$,
which is at most $-\eta(1-\eta A^2/2)\|\bar g\|^2$. With $\eta=A^{-2}$,
this upper bound becomes $-\eta\|\bar g\|^2/2$, which is strictly negative
whenever the gradient is nonzero.

Let $e_t=w_t-w^\star$. Joint realizability gives
$e_{t+1}=(I-\eta\bar G)e_t$, and hence
\[
 e_t=\left(\frac{(1-\eta A^2)^t}{2A},
           -\frac32(1-\eta)^t\right)^\top.
\]
With $\eta=A^{-2}$, the first component vanishes for $t\ge1$, and the two
prediction residuals are $-\tfrac32(1-A^{-2})^t$ and
$+\tfrac32(1-A^{-2})^t$, respectively. Squaring yields
Eq.~\eqref{app:eq:joint-transient}. Rearranging
$L_1(w_t)>L_1(w_0)=1/2$ as $(1-A^{-2})^t>2/3$ gives
Eq.~\eqref{app:eq:duration}. Finally, use
$-\log(1-A^{-2})\sim A^{-2}$ and $\kappa(\bar G)=A^2$.
\end{proof}

The first coordinate represents a direction shared by both tasks, and the
second retains information that distinguishes them. The shared direction initially
dominates the correction, and the distinguishing direction subsequently
restores task one. Degradation can therefore persist over a finite training
budget, although this model eventually learns both tasks simultaneously.

A behavior that keeps the robot in place and has already been learned can
also be perturbed without opposing nonzero residuals. Keeping $h_1,h_2$, take
$y_1=0,y_2=b>0$, $w_0=0$, and task-one weight $\alpha\in(0,1)$. Then
$g_1=0$, and the joint step is $w^+=\eta(1-\alpha)b\,h_2$, giving
\begin{equation}
 L_1(w^+)=\frac{\eta^2}{2}(1-\alpha)^2b^2(A^2-1)^2.
 \label{app:eq:stationary-contamination}
\end{equation}
When $A^2\ne1$, task one deteriorates from its original zero loss. The other
task's update changes a shared output that was already correct, corresponding
to the second-order effect in Eq.~\eqref{app:eq:joint-exact}. A common
solution $w^\star=(b/(2A),-b/2)^\top$ still exists.

These results use a local fixed-feature model and standard gradient descent.
The fixed features are an analytical approximation at a given parameter point.
Upstream vision-language modules and the action expert can change their
internal activations during fine-tuning.
The example therefore establishes a possible optimization mechanism.
For optimizers with momentum, preconditioning, or weight decay, let
$\Delta\theta_{\mathrm{actual}}$ denote the actual parameter change.
The first-order loss change is
$g_q^\top\Delta\theta_{\mathrm{actual}}$. Unpreconditioned gradient angles
cannot be substituted directly into the standard gradient descent result.

\paragraph{Expert responses cannot substitute for current gradients.}
If an unregularized expert $W_q$ is a stationary point of the fixed-feature
squared loss, then $\mathbb E_q[(W_qh-y_q)h^\top]=0$, so at the shared base,
\begin{equation}
 g_q(W_0)=-D_qG_q.
 \label{app:eq:expert-gradient}
\end{equation}
For a common input second-moment matrix $G$, further take the full factor
$S_q=S=G^{1/2}$, so that $M_q=D_qS$. Then
\begin{equation}
 \langle g_p(W_0),g_q(W_0)\rangle_F
 =\operatorname{tr}(M_pGM_q^\top),
 \qquad
 \langle M_p,M_q\rangle_F=\operatorname{tr}(M_pM_q^\top).
 \label{app:eq:different-metrics}
\end{equation}
Equation~\eqref{app:eq:expert-gradient} follows by substituting $W_0=W_q-D_q$
into the gradient. Substituting $D_pG=M_pG^{1/2}$ into the inner product then
gives Eq.~\eqref{app:eq:different-metrics}.

Take $M_1=\operatorname{diag}(10,1)$ and
$M_2=\operatorname{diag}(-20,1)$, and let each task's target be $W_qh$, with
$W_0=0$ and $W_q=M_qG^{-1/2}$. The leading response direction accounts for
$100/101$ and $400/401$ of the two experts' response energies, respectively,
and the response inner product is always $-199$. Yet at initialization of
equally weighted joint training,
\[
\begin{array}{c|rr}
G & \langle g_1,g_2\rangle_F & \langle g_1,\bar g\rangle_F\\ \hline
I & -199 & -49\\
\operatorname{diag}(10^{-4},1) & 0.98 & 0.995
\end{array}
\]
Thus, identical concentrated and opposing expert responses can yield
opposite first-order training conclusions. This counterexample establishes
that response statistics alone are insufficient. Proposition~\ref{app:prop:joint}
separately establishes transient degradation even for jointly realizable
tasks. For regularized experts, different activation distributions, or
nonlinear fine-tuning trajectories, the stationarity assumption in
Eq.~\eqref{app:eq:expert-gradient} also cannot be directly transferred.

\subsection{Two-Level On-Policy Distillation and Per-Task Returns}
\label{app:opd}

Merging error characterizes the deviation of the initialization from the
experts. OPD subsequently calibrates the student's action generation behavior
directly. This section analyzes finite-step sampling using the discrete
Gaussian transitions from the main text, retaining the time-step weights and
effective action dimension of the actual loss.

Fix task $q$. Let $T_q$ denote its teacher and $\theta$ the student parameters,
with velocity fields $v_{T_q}$ and $v_\theta$, respectively. Let $j$ index
environment decisions, $K$ be the number of sampling steps in one action
generation process, and $\ell=0,\ldots,K-1$ index flow steps.
Let $\mathsf h_j$ be the full environment history or a sufficient Markov
state, and let $o_j$ be the model conditioning determined by it. Full
histories allow the environment to include contact states, controller memory,
and fixed action prefixes. The following conditions specify the stochastic
sampling policies considered in this section:

\begin{enumerate}
\item The student and teacher execute the same task under the same initial
state distribution and environment transitions, with episode length
$1\le N\le H_q$ and the same termination rule. Finite timeouts and a common
task evaluation protocol provide this setting. The task return
$\mathcal R_q$ is the same function of the complete trajectory for both
policies and takes values in $[0,1]$, including success indicators.
\item Both policies use the same $K$-step noise schedule and initial noise
distribution. The flow time $t_\ell$ decreases along
$1=t_0>\cdots>t_K=0$, $h_\ell=t_\ell-t_{\ell+1}>0$ is the step size, and
$g_\ell>0$ is the noise coefficient at that step. Only the velocity field is
replaced at each step, as in the equal-covariance Gaussian kernels of the
main text.
\item The $d_q$ effective action coordinates used in the loss constitute the
complete stochastic sampling state. Other coordinates are fixed by functions
shared by both policies and can be included in the conditioning.
If stochastic coordinates omitted from the loss affect subsequent effective
coordinates, they must also be included in the state and KL divergence for
this analysis. Decoding, denormalization, and action execution mappings are
identical for both policies.
\end{enumerate}
These conditions specify the policies and interfaces being compared.
They do not assume that neural network training necessarily attains low error.

Let $d_q$ denote the total number of scalar coordinates in the effective
stochastic state, counting all effective coordinates of a flattened action
chunk. Let $z\in\mathbb R^{d_q}$ denote an intermediate action state and $o$
fixed observation conditioning. We use the score parameterization constructed
from the velocity field in the main text,
$s_\theta(z,t,o)=-[z+(1-t)v_\theta(z,t,o)]/t$,
which gives the reverse-time SDE drift
$v_\theta-\tfrac12g_\ell^2s_\theta$.
Taking an Euler--Maruyama step of size $h_\ell$ in decreasing time gives the
conditional mean $z-h_\ell[v_\theta-\tfrac12g_\ell^2s_\theta]$.
Define
\[
 c_\ell=1-\frac{h_\ell g_\ell^2}{2t_\ell},\qquad
 d_\ell=h_\ell\left(1+\frac{g_\ell^2(1-t_\ell)}{2t_\ell}\right).
\]
Fix the sampling step $\ell$ below and omit its index from the one-step mean
$\mu_\theta$ and transition kernel $K_\theta$. Within the effective
coordinates, the one-step conditional mean is
$\mu_\theta(z,o)=c_\ell z-d_\ell v_\theta(z,t_\ell,o)$ and the covariance is
$g_\ell^2h_\ell I_{d_q}$. The student kernel
$K_\theta(\cdot\mid z,o)$ is the Gaussian distribution with this mean and
covariance. The teacher kernel $K_{T_q}$ replaces only $v_\theta$ by
$v_{T_q}$. KL denotes Kullback--Leibler divergence with natural logarithms,
which measures the difference between two probability distributions.
Taking the expectation of the log ratio of the two Gaussian
densities under the student kernel eliminates the centered linear term,
giving
\begin{equation}
 \mathrm{KL}(K_\theta(\cdot\mid z,o)\|K_{T_q}(\cdot\mid z,o))
 =\frac{\|\mu_\theta-\mu_{T_q}\|^2}{2g_\ell^2h_\ell}
 =w_\ell\|v_\theta-v_{T_q}\|^2,\qquad
 w_\ell=\frac{d_\ell^2}{2g_\ell^2h_\ell}.
 \label{app:eq:gaussian}
\end{equation}
This $w_\ell$ matches the weight in the main text \citep{flowopd}.

Let $z_\ell^\theta$ be generated by the action sampler of the current fixed
student, with complete path
$\mathbf z^\theta=(z_0^\theta,\ldots,z_K^\theta)$ and corresponding teacher
path $\mathbf z^{T_q}$. Define the average
distillation error when this path is resampled given a history as
\begin{equation}
 \ell_q(\mathsf h)=
 \mathbb E_{\mathbf z^\theta\mid o(\mathsf h)}
 \left[\frac1K\sum_{\ell=0}^{K-1}
       \frac{w_\ell}{d_q}
       \|v_\theta(z_\ell^\theta,t_\ell,o)
          -v_{T_q}(z_\ell^\theta,t_\ell,o)\|^2\right].
 \label{app:eq:conditional-opd}
\end{equation}
Projection onto the effective coordinates is absorbed into the velocity
fields here. Further define
\begin{equation}
 \varepsilon_q^{\mathrm{self}}
 =\mathbb E_{\text{student environment episodes}}
       \left[\frac1N\sum_{j=0}^{N-1}\ell_q(\mathsf h_j)\right].
 \label{app:eq:self-opd}
\end{equation}
The quantity $\ell_q(\mathsf h)$ measures action generation error given a
history, and $\varepsilon_q^{\mathrm{self}}$ averages this error over the
environment histories that the student actually visits. Selecting
observations from student environment episodes and then resampling student
flow paths at those observations matches the two-level sampling objective in
the main text. Define $J_q(\theta)=\mathbb E_\theta[\mathcal R_q]$ and
$J_q(T_q)=\mathbb E_{T_q}[\mathcal R_q]$ as the expected task returns of the
student and teacher, respectively.

\begin{theorythm}[Per-task return bound under two-level OPD]
\label{app:thm:opd}
When the student's distillation error on its own environment states and
action paths is sufficiently small, its task performance is close to the
teacher's. Under the preceding stochastic sampling conditions, if
$\varepsilon_q^{\mathrm{self}}<\infty$, then
\begin{equation}
 |J_q(\theta)-J_q(T_q)|
 \le \min\left\{1,\sqrt{\frac{H_qKd_q}{2}
                          \varepsilon_q^{\mathrm{self}}}\right\}.
 \label{app:eq:return}
\end{equation}
\end{theorythm}

The left-hand side of Eq.~\eqref{app:eq:return} is the performance gap between
student and teacher. The right-hand side bounds it at the scale of the
square root of distillation error. A smaller
$\varepsilon_q^{\mathrm{self}}$ gives a smaller guaranteed gap.
The quantity $H_q$ is the episode horizon, while $K$ and $d_q$ correspond to
the averages over steps and coordinates in the loss. When $\mathcal R_q$ is
the success indicator, the left-hand side is the success-rate gap.

For the average multitask error
$\bar\varepsilon=\sum_q\alpha_q\varepsilon_q^{\mathrm{self}}$, we also obtain
\begin{equation}
 J_q(\theta)\ge J_q(T_q)-
 \min\left\{1,\sqrt{\frac{H_qKd_q}{2\alpha_q}\bar\varepsilon}\right\}.
 \label{app:eq:worst-task}
\end{equation}
Here, $\alpha_q>0$ and $\sum_q\alpha_q=1$ are the task weights.
For the same average distillation error, the guarantee is therefore looser
for tasks with smaller weights.

\begin{proof}[Proof]
First fix a history $\mathsf h$. Let
$\mathbb P_\theta(\mathbf z\mid\mathsf h)$ and
$\mathbb P_{T_q}(\mathbf z\mid\mathsf h)$ denote the conditional distributions
of the student's and teacher's complete action paths, respectively.
The student and teacher action paths share the same initial noise
distribution, so the log ratio of their path densities is the sum of the
log ratios of the conditional transition kernels. Taking expectations under
the student path distribution, the KL chain rule and
Eq.~\eqref{app:eq:gaussian} give
\begin{equation}
 \mathrm{KL}(\mathbb P_\theta(\mathbf z\mid\mathsf h)
                  \|\mathbb P_{T_q}(\mathbf z\mid\mathsf h))
 =\sum_{\ell=0}^{K-1}
   \mathbb E_{z_\ell^\theta\mid\mathsf h}
       [w_\ell\|v_\theta-v_{T_q}\|^2]
 =Kd_q \ell_q(\mathsf h).
 \label{app:eq:flow-path-kl}
\end{equation}
In each term, the teacher is evaluated at the student's intermediate state.

Let $\mathbb P_\theta^{\mathrm{ext}}$ and
$\mathbb P_{T_q}^{\mathrm{ext}}$ denote the extended episode distributions
containing the environment history and every complete action path. After
termination, both processes can be padded with a common absorbing process to
$H_q$ steps. The shared initial state distribution, environment transitions,
and execution mappings contribute zero conditional KL divergence. Applying
the chain rule again gives
\begin{equation}
\begin{split}
 \mathrm{KL}(\mathbb P_\theta^{\mathrm{ext}}
                  \|\mathbb P_{T_q}^{\mathrm{ext}})
 &=Kd_q\,\mathbb E_\theta\left[
                 \sum_{j=0}^{N-1}\ell_q(\mathsf h_j)\right]\\
 &\le H_qKd_q\,\mathbb E_\theta\left[
                 \frac1N\sum_{j=0}^{N-1}\ell_q(\mathsf h_j)\right]\\
 &=H_qKd_q\varepsilon_q^{\mathrm{self}}.
\end{split}
\label{app:eq:environment-kl}
\end{equation}
The inequality uses $N\le H_q$ and $\ell_q\ge0$ in each episode.
It does not require the random episode length to be independent of the error.

Use the total variation convention
$\operatorname{TV}(P,Q)=\sup_A|P(A)-Q(A)|$,
where $P,Q$ are two probability distributions and the supremum ranges over
all measurable events $A$. For any return function with values in $[0,1]$,
its expectation difference is bounded by total variation. This follows by
taking expectations under both distributions in
$\mathcal R_q=\int_0^1\mathbf 1_{\{\mathcal R_q>u\}}\,du$.
Pinsker's inequality gives
$\operatorname{TV}(P,Q)\le\sqrt{\mathrm{KL}(P\|Q)/2}$.
Substituting Eq.~\eqref{app:eq:environment-kl} and using the fact that the
return gap is at most $1$ yields Eq.~\eqref{app:eq:return}. Finally,
$\alpha_q\varepsilon_q^{\mathrm{self}}\le\bar\varepsilon$
gives Eq.~\eqref{app:eq:worst-task}.
\end{proof}

Equation~\eqref{app:eq:flow-path-kl} explains why action-level supervision is
evaluated along student paths, and Eq.~\eqref{app:eq:environment-kl} explains
why environment-level sampling uses states visited by the student.
The result directly covers discontinuous success indicators because it
compares complete stochastic trajectory distributions and does not require
the success function to be Lipschitz in the actions. The factors $K$ and $d_q$
arise because the main-text loss averages over steps and coordinates.
They do not imply that increasing the number of sampling steps or the action
dimension necessarily reduces performance. Whether the bound is numerically
informative still depends on the residual magnitude.

\paragraph{Policy and optimization scope.}
The theorem compares teachers and students satisfying the same stochastic
sampling protocol above. Equation~\eqref{app:eq:return} does not directly
carry over if deployment uses a deterministic ODE, a different noise
schedule, or different action processing. In particular, when $g_\ell=0$,
the KL formula for nondegenerate Gaussians with equal covariance does not
apply. The theorem also does not identify the empirical training loss with
the current $\varepsilon_q^{\mathrm{self}}$. States collected by previous
parameters, finite data, and the updated policy may correspond to different
distributions. It compares policies given the current error and provides no
convergence guarantee for nonconvex training.

Applying stop-gradient to the teacher does not change the numerical loss in
Eq.~\eqref{app:eq:gaussian}. However, when the student path $z^\theta$ depends
on the parameters, applying stop-gradient to the entire teacher output
removes the teacher's derivative with respect to its input. The probabilistic
bound therefore neither requires nor implies that the update used is the
full gradient of the numerical path KL divergence. Task conditioning helps
the student express different skills for different observations and
instructions. Optimization of shared parameters can still exhibit the
gradient interference analyzed in the preceding section.

\end{document}